\documentclass{article}

\usepackage{microtype}
\usepackage{graphicx}
\usepackage{subfigure}
\usepackage{booktabs} % for professional tables

\usepackage{hyperref}
\usepackage{url}
\usepackage{multirow}  % table

 \usepackage[accepted]{icml2023}

\usepackage{amsmath}
\usepackage{amssymb}
\usepackage{mathtools}
\usepackage{amsthm}

\usepackage[capitalize,noabbrev]{cleveref}

\theoremstyle{plain}

\theoremstyle{definition}

\theoremstyle{remark}

\usepackage[textsize=tiny]{todonotes}

\icmltitlerunning{Published as a Conference Paper at ICML 2023}

\begin{document}

\twocolumn[
\icmltitle{Input Perturbation Reduces  Exposure Bias in Diffusion Models}



\icmlsetsymbol{equal}{*}

\begin{icmlauthorlist}
\icmlauthor{Mang Ning}{xxx,dief}
\icmlauthor{Enver Sangineto}{dief}
\icmlauthor{Angelo Porrello}{dief}
\icmlauthor{Simone Calderara}{dief}
\icmlauthor{Rita Cucchiara}{dief}
\end{icmlauthorlist}

\icmlaffiliation{xxx}{Department of Information and Computing Science, Utrecht University, the Netherlands.}
\icmlaffiliation{dief}{Department of Engineering (DIEF), University of Modena and Reggio Emilia, Italy}
%\icmlaffiliation{sch}{School of ZZZ, Institute of WWW, Location, Country}

\icmlcorrespondingauthor{Mang Ning}{m.ning@uu.nl}
\icmlcorrespondingauthor{Enver Sangineto}{enver.sangineto@unimore.it}

\icmlkeywords{Machine Learning, ICML}

\vskip 0.3in
]



\printAffiliationsAndNotice{}  % leave blank if no need to mention equal contribution

\begin{abstract}
Denoising Diffusion Probabilistic Models  have shown an impressive generation quality  although  their long sampling chain leads to high computational costs. In this paper, we observe  that  a long sampling chain also leads to an error accumulation phenomenon, which is similar to the {\em exposure bias} problem in autoregressive text generation. Specifically, we note that there is a discrepancy between  training and testing, since the former is conditioned on the ground truth samples, while the latter is conditioned on the previously generated results. To alleviate this problem, we propose a very simple but effective training regularization, consisting in perturbing the ground truth samples to simulate the inference time prediction errors. We empirically show that, without affecting the recall and precision, the proposed input perturbation leads to a significant improvement in the sample quality while reducing both the training and the inference times.
For instance, on CelebA 64$\times$64, we achieve a new state-of-the-art FID score of 1.27,  while saving 37.5\% of the training time.
The code is available at \url{https://github.com/forever208/DDPM-IP}.
\end{abstract}

\section{Introduction}
\label{sec.Introduction}

Denoising Diffusion Probabilistic Models (DDPMs) \citep{pmlr-v37-sohl-dickstein15,DDPMs}
are a new generative paradigm which is attracting a growing interest due to its very high-quality sample generation capabilities
\citep{ADM,DBLP:conf/icml/NicholDRSMMSC22,DALL-E-2}.
Differently from most existing generative methods which synthesize a new sample in a single step, DDPMs resemble the Langevin dynamics \citep{DBLP:conf/icml/WellingT11} and the generation process is based on a sequence of denoising steps, in which a synthetic sample is created starting from pure noise and autoregressively reducing the noise component. 
In more detail, during training, a real sample $\pmb{x}_0$ is progressively destroyed in $T$ steps adding Gaussian noise ({\em forward process}). 
The sequence $\pmb{x}_0, ..., \pmb{x}_t, ..., \pmb{x}_T$ so obtained, is used to train a deep denoising autoencoder ($\mu(\cdot)$) to invert the forward process: 
$\hat{\pmb{x}}_{t-1} = \mu(\pmb{x}_t, t)$. At inference time, the generation process is autoregressive because it depends on the previously generated samples: $\hat{\pmb{x}}_{t-1} = \mu(\hat{\pmb{x}}_t, t)$
(Sec.~\ref{sec.Background}).

Despite the large success of DDPMs in different generative fields (Sec.~\ref{sec.Related}), one of the main drawbacks of these models is their very long computational time, which depends on the large number of steps $T$ required at both the training and the inference stage.
As recently emphasised in \citep{xiao2021tackling}, the fundamental reason why $T$ needs to be large is that each denoising step is assumed to be Gaussian, and this assumption holds only for small step sizes. Conversely, with larger step sizes, the prediction network ($\mu(\cdot)$) needs to solve a harder problem and it becomes progressively less accurate \citep{xiao2021tackling}.
However, in this paper, we observe that there is a second phenomenon, related to the  sampling chain,
but partially in contrast with the first,
which is the {\em accumulation} of these errors over the $T$ inference sampling steps. 
This is basically due to  the discrepancy between the training and the inference stage, in which the latter generates a sequence of samples based on the results of the previous steps, hence possibly accumulating errors.  In fact, at training time, $\mu(\cdot)$ is trained with a ground truth pair $(\pmb{x}_t, \pmb{x}_{t-1})$ and, given $\pmb{x}_t$, it learns to reconstruct $\pmb{x}_{t-1}$ ($\mu(\pmb{x}_t, t)$). However, at inference time, $\mu(\cdot)$ has no access to  the ``real'' $\pmb{x}_t$, and its prediction depends on the previously generated $\hat{\pmb{x}}_t$ ($\mu(\hat{\pmb{x}}_t, t)$). This input mismatch between $\mu(\pmb{x}_t, t)$, used during training, and $\mu(\hat{\pmb{x}}_t, t)$, used during testing, is  similar to the {\em exposure bias} problem \citep{DBLP:journals/corr/RanzatoCAZ15,DBLP:conf/emnlp/Schmidt19} shared by other autoregressive generative methods. For example, 
\citet{rennie2017self} argue that training a network  to maximize the likelihood of the next ground-truth word given the previous ground-truth word (called “Teacher-Forcing” \citep{bengio2015scheduled}) results in error accumulation  at
inference time, since the model has never been exposed to its own predictions.

In this paper, we first empirically analyze this accumulation error phenomenon. For instance, we show that a standard DDPM \citep{ADM}, trained with $T$ steps, can generate {\em better} results using a number of inference steps $T' < T$ (Sec.~\ref{sec.main results}). 
A similar phenomenon was also observed by \citet{IDDPM}, but the authors did not provide an explanation for that. We believe that the reason for this apparently contrasting result is that while,  on the one hand, longer chains can better satisfy the Gaussian assumption in the reverse diffusion process,  on the other hand, they lead to a larger accumulation of errors.

Second, in order to alleviate the exposure bias problem, we propose a surprisingly simple yet very effective method, which consists in explicitly modelling the prediction error during training. Specifically, at training time, we perturb $\pmb{x}_t$ and we feed $\mu(\cdot)$ with a noisy version of $\pmb{x}_t$, this way simulating the training-inference discrepancy, and forcing the learned network to take into account possible inference-time prediction errors. Note that our perturbation is different from the content-destroying forward process, because the new noise is {\em not} used in the ground truth prediction {\em target}  (Sec.~\ref{sec.Discussion}).
The proposed method is a training regularization which forces the  network to smooth its prediction function: to solve the proposed task, two spatially close  points $\pmb{x}_1$ and $\pmb{x}_2$ should lead to similar predictions $\mu(\pmb{x}_1, t)$ and $\mu(\pmb{x}_2, t)$. 
This  regularization approach is similar to  Mixup \cite{mixup} and the  Vicinal Risk Minimization (VRM) principle \citep{DBLP:conf/nips/ChapelleWBV00}, where a neighborhood
around each sample in the training data is defined and then used to perturb that sample keeping fixed its target class label.

Third, we propose alternative solutions to the exposure bias problem for diffusion models, in which, rather than using input perturbation, we obtain a 
smoother prediction function $\mu(\cdot)$ by  explicitly encouraging  $\mu(\cdot)$ to be Lipschitz continuous (Sec.~\ref{sec.Lipschitz}). The rationale behind this is that a Lipschitz continuous function  $\mu(\cdot)$  generates small prediction differences between neighbouring points in its domain, leading to a DDPM which is more robust to the inference-time errors. 

Finally, we empirically analyse all the proposed solutions and we show that, despite being all effective for improving the final generation quality, input perturbation is 
both more efficient and more effective than
the explicit minimization of the Lipschitz constant in DDPMs (Sec.~\ref{sec.Lipschitz-results}).
Moreover, directly perturbing the network input at training time has {\em no additional training overhead} and this solution is very easy to be reproduced and plugged into existing DDPM frameworks: 
 it can be obtained 
with just two lines of code without any change in the network architecture or the loss function.
 We call our method  {\em Denoising Diffusion Probabilistic Models with Input Perturbation (DDPM-IP)} and we show that it can significantly improve the generation quality of state-of-the-art DDPMs \citep{ADM,DBLP:conf/iclr/SongME21} and speed up the inference-time sampling. For instance,
on the CIFAR10 \citep{cifar10}, the ImageNet 32$\times$32 \citep{imagenet32}, the LSUN  64$\times$64 \citep{LSUN} 
and the FFHQ 128$\times$128 \cite{karras2019style}
datasets, DDPM-IP, with only 80 sampling steps, generates lower FID scores than the state-of-the-art ADM \cite{ADM} with 1,000 steps,  corresponding to a more than $12.5 \times$ sampling acceleration.
 
In summary, our contributions are:

\begin{itemize}
    \item We  show that there is an exposure bias problem in DDPMs which has not been investigated so far. 
    \item 
    To alleviate this problem,  we propose different regularization methods whose common goal is to smooth the prediction function, and we specifically   suggest input perturbation (DDPM-IP) as the best and the simplest of such solutions.
    \item Using  common  benchmarks, we show that DDPM-IP can significantly improve the generation quality and drastically speed up both training and inference.
\end{itemize}

\section{Related Work}
\label{sec.Related}

Diffusion models were introduced by \citet{pmlr-v37-sohl-dickstein15} and later improved in
\citep{NCSN, DDPMs, VPVE, IDDPM}. More recently,
\citet{ADM} have shown that DDPMs can yield higher-quality images than Generative Adversarial Networks (GANs) \citep{goodfellow2014generative, BigGAN}. Similarly to GANs, the generation process in DDPMs can be both unconditional and conditioned. For instance, GLIDE \citep{DBLP:conf/icml/NicholDRSMMSC22} learns to generate images according to an input textual sentence. 
Differently from GLIDE, where the diffusion model is defined on the image space, DALL$\cdot$E-2 (\citet{DALL-E-2}) uses a DDPM to learn a prior distribution on the CLIP \citep{radford2021learning} space. 
Text-to-image generation is explored also in Stable Diffusion \citep{stableDiffusion} and Imagen \citep{Imagen}.
Apart from images, DDPMs can also be used with categorical distributions \citep{DBLP:conf/nips/HoogeboomNJFW21,DBLP:journals/corr/abs-2111-14822}, in an audio domain 
\citep{DBLP:conf/ismir/MittalEHS21,DBLP:conf/iclr/ChenZZWNC21}, in time series forecasting \citep{rasul2021autoregressive} and in other generative tasks  \citep{survey,survey2}. Differently from previous work, our goal is not to propose an application-specific prediction network, but rather to investigate the training-testing discrepancy of the DDPMs and propose a solution which can be used in different application fields and jointly with different denoising architectures.

Accelerating the DDPM training or reducing the number of sampling steps $T$ (Sec.~\ref{sec.Introduction}) have been thoroughly investigated due to their practical implications. For instance, \citet{DBLP:conf/iclr/SongME21} 
propose Denoising Diffusion
Implicit Models (DDIMs), based on
a non-Markovian diffusion process,  which can use a number of inference sampling steps smaller than those used at training time,  without retraining the network. 
 \citet{DBLP:conf/iclr/SalimansH22} propose to distil the prediction network into new networks which progressively reduce the number of sampling steps. However, the disadvantage is the need of training multiple networks.  
\citet{stableDiffusion} speed up sampling by splitting the process into a compression stage and a generation stage, and applying the DDPM on the compressed (latent) space.
\citet{DBLP:conf/iclr/HoogeboomGBPBS22} present an order-agnostic DDPM, inspired by XLNet \citep{DBLP:conf/nips/YangDYCSL19}, in which the sequence $\pmb{x}_0, ..., \pmb{x}_T$ is randomly permuted at training time, leading to a partially parallelized  sampling process.
 \citet{DBLP:conf/iclr/ChenZZWNC21} found that, instead of conditioning the prediction network ($\mu(\cdot)$) on a discrete diffusion step $t$, it is beneficial to condition $\mu(\cdot)$ on a continuous noise level. Similarly, \citet{kong2021fast} introduce continuous diffusion steps, resulting in a unified framework for fast sampling.  
 In order to use larger size  sampling steps and a non-Gaussian reverse process (Sec.~\ref{sec.Introduction})
 \citet{xiao2021tackling} include an adversarial loss in DDPMs and propose Denoising Diffusion GANs.
 \citet{karras2022elucidating} suggest using Heun's second-order deterministic sampling method, leading to high quality results and fast sampling. \citet{xu2022poisson} accelerate the generation process of continuous normalizing flow using a Poisson flow generative model. Our approach is orthogonal to these previous works, and it can potentially be used jointly with most of them.

\section{Background}
\label{sec.Background}

Without loss of generality, we assume an image domain and we focus on DDPMs which  define a diffusion process on the input space.  Following \citep{IDDPM,ADM}, we assume that each pixel value is linearly scaled into $[-1, 1]$. 
Given a sample $\pmb{x}_0$ from the data distribution $q(\pmb{x}_0)$ 
and a prefixed noise schedule ($\beta_1, ..., \beta_T$),
a DDPM defines the forward process as a Markov chain which starts from a real image $\pmb{x}_0 \sim q(\pmb{x}_0)$ and iteratively adds Gaussian noise for $T$ diffusion steps:

\begin{equation}
\label{eq.forward-process-1}
q(\pmb{x}_t | \pmb{x}_{t-1}) = {\cal N} (\pmb{x}_t; \sqrt{1- \beta_t} \pmb{x}_{t-1}, \beta_t \pmb{I}),
\end{equation}

\begin{equation}
\label{eq.forward-process-2}
q(\pmb{x}_{1:T} | \pmb{x}_0) = \prod_{t=1}^T q(\pmb{x}_t | \pmb{x}_{t-1}),
\end{equation}

\noindent
until obtaining a completely noisy image $\pmb{x}_T \sim {\cal N} (\pmb{0}, \pmb{I})$.
On the other hand, the {\em reverse process} is defined by  transition probabilities parameterized by $\pmb{\theta}$:

\begin{equation}
\label{eq.backward-process-1}
p_{\pmb{\theta}}(\pmb{x}_{t-1} | \pmb{x}_t) = {\cal N} (\pmb{x}_{t-1}; \mu_{\pmb{\theta}}(\pmb{x}_t, t), \sigma_t \pmb{I}),
\end{equation}

% \begin{equation}
% \label{eq.backward-process-2}
% p_{\pmb{\theta}}(\pmb{x}_{0:T} ) = p(\pmb{x}_T) \prod_{t=1}^T p_{\pmb{\theta}}(\pmb{x}_{t-1} | \pmb{x}_t),
% \end{equation}

\noindent
where $\sigma_t = \frac{1-\bar{\alpha}_{t-1}}{1-\bar{\alpha}_t} \beta_t$ with $\bar{\alpha}_t = \prod_{i=1}^t \alpha_i$ and $\alpha_i = 1 - \beta_i$.
Given $\pmb{x}_0$, $\pmb{x}_t$ can be obtained  \citep{DDPMs} by: 

\begin{equation}
\label{eq.x-t}
\pmb{x}_t = \sqrt{\bar{\alpha}_t} \pmb{x}_0 +   \sqrt{1 - \bar{\alpha}_t} \pmb{\epsilon},
\end{equation}

\noindent
 where $\pmb{\epsilon}$ is a noise vector 
($\pmb{\epsilon} \sim {\cal N} (\pmb{0}, \pmb{I})$). 
Instead of predicting the mean of the forward process posterior (i.e., $\hat{\pmb{x}}_{t-1} = \mu_{\pmb{\theta}}(\pmb{x}_t, t)$),
\citet{DDPMs} propose to use a network $\pmb{\epsilon}_{\pmb{\theta}}(\cdot)$ which predicts the noise vector ($\pmb{\epsilon}$). Using $\pmb{\epsilon}_{\pmb{\theta}}(\cdot)$ and a simple $L_2$ loss function, the training objective becomes:

\begin{equation}
\label{eq.L2-loss}
L(\pmb{\theta}) = \mathbb{E}_{\pmb{x}_0 \sim q(\pmb{x}_0), \pmb{\epsilon} \sim {\cal N} (\pmb{0}, \pmb{I}), t \sim \mathbb{U}(\{1,...,T\})} [||\pmb{\epsilon} - 
\pmb{\epsilon}_{\pmb{\theta}} (\pmb{x}_t, t)  ||^2].
\end{equation}

Note that, in Eq.~\ref{eq.L2-loss}, $\pmb{x}_t$ and $\pmb{\epsilon}$ are ground-truth terms, while $\pmb{\epsilon}_{\pmb{\theta}} (\pmb{x}_t, t)$ is the network prediction. Using Eq.~\ref{eq.L2-loss}, the training and the sampling algorithms are described in Alg.~\ref{alg:1}-\ref{alg:2}, respectively.

\begin{algorithm}[h]
   \caption{DDPM Standard Training}
   \label{alg:1}
\begin{algorithmic}[1]
   \REPEAT
        \STATE $\pmb{x}_0 \sim q(\pmb{x}_0)$, $t\sim \mathbb{U}(\{1,...,T\})$, $\pmb{\epsilon} \sim {\cal N} (\pmb{0}, \pmb{I})$
        \STATE compute  $\pmb{x}_t$ using Eq.~\ref{eq.x-t}        
        \STATE $\text{take a gradient descent step on} \:
         \nabla_{\pmb{\theta}} ||\pmb{\epsilon} - \pmb{\epsilon}_{\pmb{\theta}} (\pmb{x}_t, t)  ||^2$ \label{line.1}
%        \phantom{x}\hspace{5ex} \nabla_{\pmb{\theta}} ||\pmb{\epsilon} - \pmb{\epsilon}_{\pmb{\theta}} (\pmb{x}_t, t)  ||^2$ \label{line.1}
        \UNTIL {converged}
\end{algorithmic}
\end{algorithm}

\begin{algorithm}[h]
   \caption{DDPM Standard Sampling}
   \label{alg:2}
\begin{algorithmic}[1]
        \STATE $\hat{\pmb{x}}_T \sim  {\cal N} (\pmb{0}, \pmb{I})$
        \FOR{$t := T, ..., 1$}
        \STATE if $t > 1$ then $\pmb{z} \sim {\cal N} (\pmb{0}, \pmb{I})$, else $\pmb{z} = \pmb{0}$ 
        \STATE $\hat{\pmb{x}}_{t-1} = \frac{1}{\sqrt{\alpha_t}} (\hat{\pmb{x}}_t - \frac{1-\alpha_t}{\sqrt{1-\bar{\alpha}_t}} \pmb{\epsilon}_{\pmb{\theta}} (\hat{\pmb{x}}_t, t)) + \sigma_t \pmb{z}$ \label{line.2}
        \ENDFOR
        \STATE \textbf{return} $\hat{\pmb{x}}_{0}$
    \end{algorithmic}
\end{algorithm}

\section{Exposure Bias Problem in Diffusion Models}
\label{sec.Exposure}

Comparing line 4 of Alg.~\ref{alg:1} with line 4 of Alg.~\ref{alg:2}, we note that the inputs of the prediction network 
$\pmb{\epsilon}_{\pmb{\theta}}(\cdot)$ are different between the training  and the inference phase. Concretely, at training time, standard DDPMs use 
$\pmb{\epsilon}_{\pmb{\theta}} (\pmb{x}_t, t)$, where $\pmb{x}_t$ is a {\em ground truth} sample (Eq.~\ref{eq.x-t}). In contrast, at inference time, they use $\pmb{\epsilon}_{\pmb{\theta}} (\hat{\pmb{x}}_t, t))$, where $\hat{\pmb{x}}_t$ is computed based on the output of $\pmb{\epsilon}_{\pmb{\theta}}(\cdot)$ at the previous sampling step t$+$1. As mentioned in Sec.~\ref{sec.Introduction}, this leads to a training-inference discrepancy, which is similar to the exposure bias problem observed, e.g., in text generation models, in  which the training generation is conditioned on a ground-truth sentence, while the testing generation is conditioned on the previously generated words \citep{DBLP:journals/corr/RanzatoCAZ15,DBLP:conf/emnlp/Schmidt19,rennie2017self,bengio2015scheduled}. 
In order to quantify 
the error {\em accumulation} with respect to the number of inference sampling steps, 
we use a simple experiment in which we 
start from a (randomly selected)  real image $\pmb{x}_{0}$,  we compute $\pmb{x}_t$
using Eq.~\ref{eq.x-t}, and then apply the reverse process (Alg.~\ref{alg:2}) {\em starting from} $\pmb{x}_t$ instead of a random $\pmb{x}_T$. This way, when $t$ is small enough, the network should be able to ``recover'' the path to $\pmb{x}_{0}$
(the denoising task is easier).  We quantify the total error accumulated in $t$ reverse diffusion steps by comparing the difference between the ground truth distribution $q(\pmb{x}_0)$ and the predicted distribution $q(\hat{\pmb{x}}_{0})$
using the FID scores in Tab.~\ref{stochastic exposure bias}. The experiment was done using 
 ADM \citep{ADM} (trained  with $T=1,000$)  and   ImageNet 32$\times$32,
 and we compute the FID scores using 50k samples.
 Tab.~\ref{stochastic exposure bias} (first row)  shows that {\em the longer the reverse process, the higher the FID scores}, indicating the existence of an error accumulation which is larger with larger values of $t$.
 In Appendix~\ref{exposure bias}, we repeat this experiment using deterministic sampling, which quantifies the error accumulation removing the randomness from the sampling process.

\begin{table}[h]
\caption{
An empirical estimate of the exposure bias on ImageNet 32$\times$32.}
\label{stochastic exposure bias}
\begin{center}
\begin{tabular}{@{}llllll@{}}
\toprule
\multirow{2}{*}{Model} & \multicolumn{5}{l}{Number of reverse diffusion steps} \\ \cmidrule(l){2-6} 
 & 100 & 300 & 500 & 700 & 1,000 \\ \midrule
ADM & 0.983 & 1.808 & 2.587 & 3.105 & 3.544 \\
ADM-IP (ours) & 0.972 & 1.594 & 2.198 & 2.539 & 2.742 \\ \bottomrule
\end{tabular}
\end{center}
\end{table}

Finally, in 
 Tab.~\ref{DDPM results}
we will report the FID scores of ADM on different datasets, which show that most of the best results are obtained in the range from 100 to 300 sampling steps, despite all the models have been trained with 1,000 diffusion steps. These results 
confirm previous similar observations \cite{IDDPM}, and we believe that the reason for this 
apparently counterintuitive phenomenon, in which fewer sampling  steps lead to a better generation quality, is due to the exposure bias problem. Indeed, 
while more sampling steps correspond to a reverse process which can be more easily approximated with a Gaussian distribution (Sec.~\ref{sec.Introduction}), longer sampling trajectories produce a larger accumulation of the prediction errors. Hence,  the range $[100, 300]$ leads to a better generation quality because it presumably trades off these two opposing aspects.

\section{Method}
\label{sec.Method}

\subsection{Regularization with Input Perturbation}
\label{subsec.IP alleviate EB}

The solution we propose to alleviate the exposure bias problem is very simple: we explicitly model the prediction error using a Gaussian input perturbation at training time. More specifically, we assume  that the error of the prediction network in the reverse process at time $t+1$ is normally distributed with respect to the ground-truth input $\pmb{x}_t$ (see Sec.~\ref{subsec.gamma choice}). This is simulated using a second, dedicated  random noise vector $\pmb{\xi} \sim {\cal N} (\pmb{0}, \pmb{I})$, using which, we create a  perturbed version ($\pmb{y}_t$) of $\pmb{x}_t$:

\begin{equation}
\label{eq.y-t}
\pmb{y}_t = \sqrt{\bar{\alpha}_t} \pmb{x}_0 +   \sqrt{1 - \bar{\alpha}_t} (\pmb{\epsilon} + \gamma_t \pmb{\xi}).
\end{equation}

For simplicity, we use a uniform noise schedule for $\pmb{\xi}$ by setting $\gamma_0 = ... = \gamma_T = \gamma$. In fact, 
although selecting the best noise schedule ($\beta_1, ..., \beta_T$) in DDPMs is  usually very important  to get high-quality results \citep{DDPMs,DBLP:conf/iclr/ChenZZWNC21},
 it is nevertheless an expensive hyperparameter tuning operation \citep{DBLP:conf/iclr/ChenZZWNC21}. Therefore, to avoid adding a second noise schedule ($\gamma_0, ..., \gamma_T$) to the training procedure, we opted for a simpler (although most likely sub-optimal) solution, in which $\gamma_t$ does not vary depending on $t$ (more details in Sec.~\ref{subsec.gamma choice}). In Alg.~\ref{alg:4} we show the proposed training algorithm, in which $\pmb{x}_t$ is replaced by $\pmb{y}_t$. In contrast, at inference time, we use Alg.~\ref{alg:2} without any change.

\begin{algorithm}[h]
   \caption{DDPM-IP: Training with input perturbation}
   \label{alg:4}
   \begin{algorithmic}[1]
        \REPEAT
        \STATE $\pmb{x}_0 \sim q(\pmb{x}_0)$, $t\sim \mathbb{U}(\{1,...,T\})$
        \STATE $\pmb{\epsilon} \sim {\cal N} (\pmb{0}, \pmb{I})$, $\pmb{\xi} \sim {\cal N} (\pmb{0}, \pmb{I})$
        \STATE compute  $\pmb{y}_t$ using Eq.~\ref{eq.y-t}        
        \STATE $\text{take a gradient descent step on} \: \nabla_{\pmb{\theta}} ||\pmb{\epsilon} - \pmb{\epsilon}_{\pmb{\theta}} (\pmb{y}_t, t)  ||^2$          
        %$ \phantom{x}\hspace{5ex} \nabla_{\pmb{\theta}} ||\pmb{\epsilon} - \pmb{\epsilon}_{\pmb{\theta}} (\pmb{y}_t, t)  ||^2$ 
         \label{line.4}
        \UNTIL {converged}
    \end{algorithmic}
\end{algorithm}

\subsection{Discussion}
\label{sec.Discussion}

In this section, we analyze the difference between Alg.~\ref{alg:4} and Alg.~\ref{alg:1}. Specifically,  in line 5 of Alg.~\ref{alg:4}, we use $\pmb{y}_t$ as the input of the prediction network $\pmb{\epsilon}_{\pmb{\theta}}(\cdot)$ but we keep using 
$\pmb{\epsilon}$ as the regression target. 
In other words, the new  noise term ($\pmb{\xi}$) we introduce is used {\em asymmetrically}, because it is applied to the input but {\em not} to the prediction target ($\pmb{\epsilon}$). 
For this reason, Alg.~\ref{alg:4} is {\em not} equivalent to 
 choose a different  value of $\pmb{\epsilon}$   in  Alg.~\ref{alg:1}, where $\pmb{\epsilon}$ is instead used {\em symmetrically}
 both in the forward process (Eq.~\ref{eq.x-t}) and as the target of the prediction network  (line 4 of Alg.~\ref{alg:1}).

\begin{figure}[ht]
\vskip 0.2in
\begin{center}
\centerline{\includegraphics[width=\columnwidth]{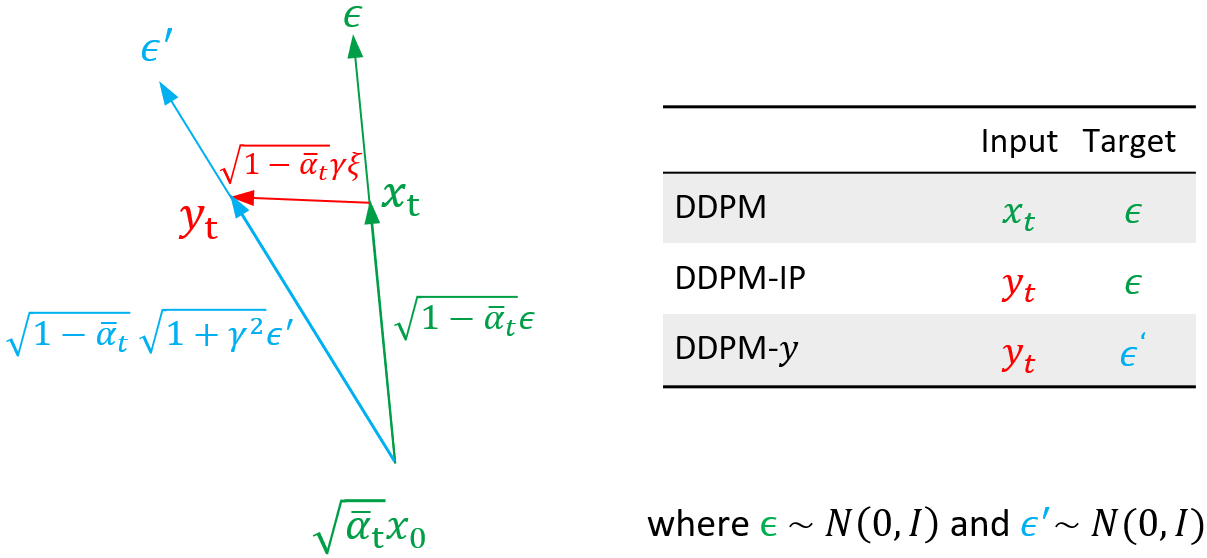}}
\caption{The inputs and the prediction targets are different in vanilla DDPM, DDPM-IP and DDPM-$y$.
}
\label{IP and noise factor}
\end{center}
\vskip -0.2in
\end{figure}

 This difference is schematically illustrated in Fig.~\ref{IP and noise factor}, where, for both
 Alg.~\ref{alg:1} (i.e., DDPM) and Alg.~\ref{alg:4} (DDPM-IP),
 we show the corresponding pairs of input and target vectors of the prediction network (respectively, $(\pmb{x}_t, \pmb{\epsilon})$
 and $(\pmb{y}_t, \pmb{\epsilon})$). In the same figure, we also show a second version of Alg.~\ref{alg:1}
(called DDPM-$y$), 
 where we use the standard training protocol (Alg.~\ref{alg:1}) but change the noise variance in order to adhere to the same distribution generating $\pmb{y}_t$. In fact, it can be easy shown  that 
 $\pmb{y}_t$ in Alg.~\ref{alg:4} is generated using the following distribution 
 (see Appendix~\ref{sec:y-distributio-proof} for a proof):

\begin{equation}
\label{eq.y-distribution}
q(\pmb{y}_t | \pmb{x}_0) =   {\cal N} (\pmb{y}_t; \sqrt{\bar{\alpha}_t} \pmb{x}_0, (1 - \bar{\alpha}_t) (1 + \gamma^2) \pmb{I}).
\end{equation}

\noindent
Hence, we can obtain {\em the same input noise distribution} of Alg.~\ref{alg:4} in Alg.~\ref{alg:1} using 
$\pmb{\epsilon^\prime} \sim {\cal N} (\pmb{0}, \pmb{I})$ and:

\begin{equation}
\label{eq.y-t-new-noise}
\pmb{y}_t = \sqrt{\bar{\alpha}_t} \pmb{x}_0 +   \sqrt{1 - \bar{\alpha}_t} \sqrt{1+\gamma^{2}}\pmb{\epsilon^\prime}.
\end{equation}

We call DDPM-$y$ the version of Alg.~\ref{alg:1} with this new noise distribution. 
DDPM-$y$ is  obtained from Alg.~\ref{alg:1}  
using  Eq.~\ref{eq.y-t-new-noise} in line 3
and replacing
$\pmb{x}_t$ with $\pmb{y}_t$ and $\pmb{\epsilon}$ with $\pmb{\epsilon^\prime}$ in line 4.
However, note that, for a given $\pmb{y}_t$, if $\pmb{\xi} \neq \pmb{0}$, then 
$\pmb{\epsilon} \neq \pmb{\epsilon^\prime}$ (see Fig.~\ref{IP and noise factor}),
thus,  DDPM-IP and DDPM-$y$  share the same input to $\pmb{\epsilon}_{\pmb{\theta}}(\cdot)$, but they use different targets. 
In Appendix~\ref{sec:Ablation study},
we empirically show that DDPM-$y$ is even worse than the standard DDPM.

Intuitively, the proposed training protocol,  DDPM-IP, decouples the noise vector
$\pmb{\epsilon^\prime}$ actually
generating $\pmb{y}_t$ from the ground truth target vector $\pmb{\epsilon}$ which is asked  to be predicted by $\pmb{\epsilon}_{\pmb{\theta}}(\cdot)$. 
In order to solve this problem, $\pmb{\epsilon}_{\pmb{\theta}}(\cdot)$ needs to {\em smooth} its prediction function, reducing the difference between $\pmb{\epsilon}_{\pmb{\theta}} (\pmb{x}_t, t)$ and $\pmb{\epsilon}_{\pmb{\theta}} (\pmb{y}_t, t)$, and this leads to a training regularization which is similar to VRM (Sec.~\ref{sec.Introduction}).

% Finally, in Appendix~\ref{sec:Score function learned by DDPM-IP} we show the relation with the score functions respectively learned by DDPM and DDPM-IP.

\subsection{Estimating the Prediction Error}
\label{subsec.gamma choice}

In this section, we analyze the actual prediction error of $\pmb{\epsilon}_{\pmb{\theta}}(\cdot)$ and we use this analysis to choose the value of $\gamma$ in Eq.~\ref{eq.y-t}. Analogously to Sec.~\ref{sec.Exposure}, we use ADM,  {\em trained using the standard algorithm Alg.~\ref{alg:1}} and two datasets: CIFAR10 and ImageNet 32$\times$32.
At testing time, for a given $t$ and 
$\hat{\pmb{\epsilon}} = \pmb{\epsilon}_{\pmb{\theta}} (\hat{\pmb{x}}_t, t)$, we 
replace $\pmb{\epsilon}$ with $\hat{\pmb{\epsilon}}$ in Eq.~\ref{eq.x-t} and we 
compute the predicted $\hat{\pmb{x}}_{0}$.
%\begin{equation}
%\label{eq.x-0-predicted}
%\hat{\pmb{x}}_{0} = \frac{\pmb{x}_t - \sqrt{1 - \bar{\alpha}_t} \hat{\pmb{\epsilon}}}{\sqrt{\bar{\alpha}_t}}
%\end{equation}
Finally, the prediction error at time $t$ is 
$\pmb{e}_{t}= \hat{\pmb{x}}_{0}-\pmb{x}_{0}$.
Note that using $\hat{\pmb{x}}_{0}$ and $\pmb{x}_{0}$ to estimate the error instead of  comparing 
$\hat{\pmb{x}}_{t}$ and $\pmb{x}_{t}$, has the advantage that the former is independent of scaling factors ($\sqrt{1 - \bar{\alpha}_t}$) and, thus, it makes the statistical analysis easier.
Using different values of $t$, uniformly selected in $\{1, ..., T\}$, we empirically verified that, for a given $t$, 
$\pmb{e}_{t}$ is normally distributed: $\pmb{e}_{t} \sim {\cal N} (\pmb{0}, \nu_{t}^{2}\pmb{I})$, with standard deviation $\nu_{t}$ (see Appendix~\ref{sec:gaussian prediction error}).

In Fig.~\ref{gaussian prediction error} we plot the value of $\nu_{t}$ with respect to $t$. The two curves corresponding to the two datasets are surprisingly close to each other. In principle, we could use this empirical analysis and set $\gamma_t = \nu_t$
in Eq.~\ref{eq.y-t}. In this way,  
when we perturb the input to $\pmb{\epsilon}_{\pmb{\theta}}(\cdot)$, we empirically imitate its actual prediction error which is the base of the exposure bias problem.
However, this choice would require a two-step training: first, using  Alg.~\ref{alg:1} to train the base model and empirically estimate $\nu_{t}$ for different $t$. Then, using Alg.~\ref{alg:4} with the estimated $\gamma_t$ schedule to retrain the model from scratch. To avoid this and make the whole procedure as simple as possible, we simply use a constant value $\gamma$, independently of $t$. This value was empirically set 
using
a grid search 
on both CIFAR10 and ImageNet 32$\times$32  on a small range of values covering the last half of the sampling trajectory. 
Specifically, we investigated the range  $\nu_{t} \in [0, \mathbb{E}_t [\nu_{t}]] = [0, 0.2]$ (see Fig.~\ref{gaussian prediction error}), which was chosen following \citet{karras2022elucidating}, who showed that the last part of the inference trajectory has usually the largest impact on the Diffusion Model performance.
%In Sec.~\ref{sec.main results} and 
We finally set 
 $\gamma = 0.1$ and, 
in the rest of this paper, we {\em always} use a constant $\gamma = 0.1$, {\em regardless of the dataset and the baseline DDPM}. Although a DDPM-specific $\gamma$ value would most likely lead to  better quality results, we prefer to emphasise the ease of use of our proposal {\em which does not depend on any other hyperparameter}.

% \begin{figure}[ht]
% \vskip 0.2in
% \begin{center}
% \centerline{\includegraphics[width=0.75\columnwidth]{images/gaussian_error_by_x_0_imgnet0.20_cifar0.19.png}}
% \caption{The inference time standard deviation $\nu_{t}$ of the prediction error of a pre-trained network with respect to the sampling step $t$. The mean of the blue and the orange curve is 0.20 and  0.19, respectively.}
% \label{gaussian prediction error}
% \end{center}
% \vskip -0.2in
% \end{figure}

\begin{figure}[ht]
% \vskip 0.2in
\begin{center}
\centerline{\includegraphics[width=0.75\columnwidth]{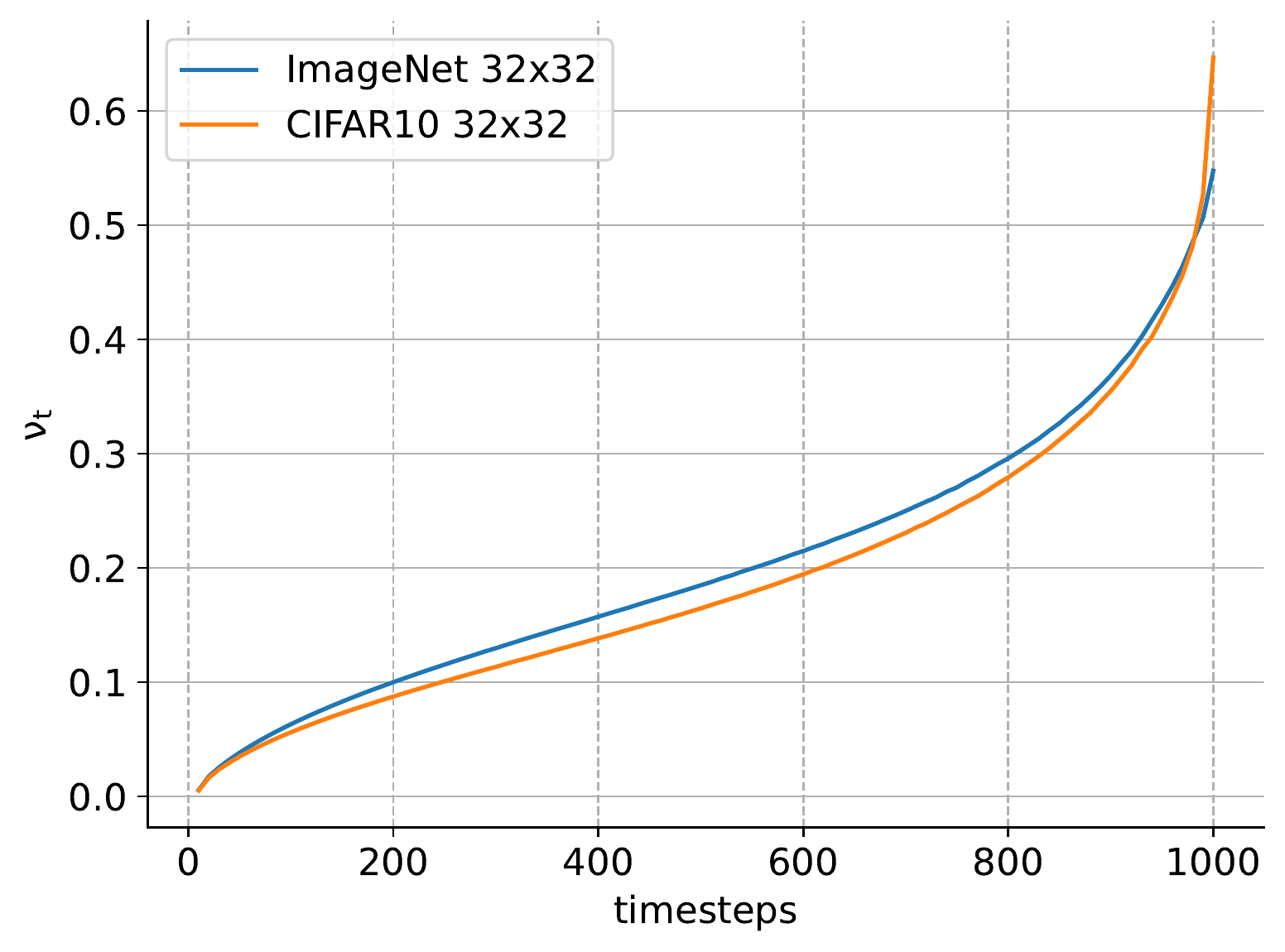}}
\caption{The inference time standard deviation $\nu_{t}$ of the prediction error of a pre-trained network with respect to the sampling step $t$. The mean of the blue and the orange curve is 0.20 and  0.19, respectively.}
\label{gaussian prediction error}
\end{center}
% \vskip -0.2in
\end{figure}

\subsection{Regularization based on Lipschitz Continuous Functions}
\label{sec.Lipschitz}
In this section, we propose two alternative solutions to the exposure bias problem which can help to better investigate the phenomenon. The goal is the same as in Sec.~\ref{subsec.IP alleviate EB}, i.e., we want to smooth the prediction function $\pmb{\epsilon}_{\pmb{\theta}}(\pmb{x}_t, t)$ to make it more robust with respect to local variations of $\pmb{x}_t$ which are due to the inference-time prediction errors.
To do so, instead of using input perturbation, we explicitly encourage $\pmb{\epsilon}_{\pmb{\theta}}(\cdot)$ to be Lipschitz continuous, i.e. to satisfy:

\begin{equation}
    \label{eq.Lipschitz-continuous}
    || \pmb{\epsilon}_{\pmb{\theta}} (\pmb{x}, t) - \pmb{\epsilon}_{\pmb{\theta}} (\pmb{y}, t) || \leq K || \pmb{x} - \pmb{y} ||,
    \: \:  \:  \forall (\pmb{x}, \pmb{y})
\end{equation}

\noindent
for a small constant $K$. We implement this idea using two standard Lipschitz constant minimization methods: {\em gradient penalty} \cite{ContractiveAutoencoder,WGAN-GP} and {\em weight decay} \cite{WeightDecay,SpectralNorm2018}.
In both cases we do {\em not} perturb the  input of $\pmb{\epsilon}_{\pmb{\theta}}(\cdot)$, and we use the original training algorithm (Alg.~\ref{alg:1}), with the only difference being the loss function used in line 4, where the $L_2$ loss is used jointly with a regularization term described below.

{\bf Gradient penalty.} In this case, the regularization is based on the Frobenius norm of the Jacobian matrix \citep{ContractiveAutoencoder,DeepLearning},  and the final loss is:

\begin{equation}
    \label{eq.Gradient-penalty}
    L_{GP} (\pmb{\theta}) = ||\pmb{\epsilon} - \pmb{\epsilon}_{\pmb{\theta}} (\pmb{x}_t, t)  ||^2 + 
    \lambda_{GP}  \left\| \frac{\partial \pmb{\epsilon}_{\pmb{\theta}} (\pmb{x}_t, t)}{\partial \pmb{x}} \right\|^2_F,
%    \lambda_{GP}  || \nabla_{\pmb{x}} [ \pmb{\epsilon}_{\pmb{\theta}} (\pmb{x}_t, t) ] ||^2,
\end{equation}

\noindent
where $\lambda_{GP}$ is the weight of the gradient penalty term. However, a gradient penalty regularization is very slow \cite{SpectralNorm2017} because it involves one forward and two backward passes for each training step. 

{\bf Weight decay.} 
As shown in \citep{LipsWeightDecay}, Lipschitz continuity can also be encouraged using a weight decay regularization (see Appendix~\ref{Lips and weight decay} for more details). In this case,  the final loss is:

\begin{equation}
    \label{eq.Weight-decay}
    L_{WD} (\pmb{\theta}) = ||\pmb{\epsilon} - \pmb{\epsilon}_{\pmb{\theta}} (\pmb{x}_t, t)  ||^2 + 
    \lambda_{WD}   || \pmb{\theta} ||^2,
\end{equation}

\noindent
where $\lambda_{WD}$ is the weight of the regularization term.

\section{Results}
\label{sec.results}

In  this section, we evaluate the generation quality of the proposed solutions and we compare them with state-of-the-art DDPMs. We use
unconditional image generation tasks on different datasets and  standard metrics:  the Fréchet Inception Distance (FID) \citep{FID} and the Spatial Fréchet Inception Distance (sFID) \citep{sFID}. As a variant of FID, sFID uses spatial features rather than the standard pooled features to better capture spatial relationships, rewarding image distributions with a coherent high-level structure. 
As mentioned in Sec.~\ref{subsec.gamma choice}, in {\bf all} our experiments we use $\gamma = 0.1$ {\em without any dataset or baseline specific tuning of our only hyperparameter}.

\subsection{Evaluation of the Different Proposed Solutions}
\label{sec.Lipschitz-results}

In this section, we empirically compare to each other the three regularization methods
proposed in Sec.~\ref{sec.Method} to alleviate the
exposure bias problem.
For all three approaches, we use the state-of-the-art diffusion model ADM \cite{ADM} (without classifier guidance) as the baseline, and we call: (1) ``ADM-IP'' the version of ADM trained using  Alg.~\ref{alg:4}, 
(2) ``ADM-GP'' the version of ADM trained using the gradient penalty, and (3) ``ADM-WD'' for the  weight decay 
(Sec.~\ref{sec.Lipschitz}).
%In ADM-IP we use $\gamma = 0.1$ (see Sec.~\ref{subsec.gamma choice}), while w
We use 
$\lambda_{GP}=1e-6$ and $\lambda_{WD}=0.03$ as the loss weights for ADM-GP and  ADM-WD, respectively.

For this experiment, we use CIFAR10 because ADM-GP is too time-consuming to be trained on larger datasets. The results in Tab.~\ref{FID Ablation} show that all three models outperform the baseline in image quality, demonstrating the effectiveness of
smoothing the prediction function using the proposed regularization methods. However, 
training ADM-GP is too slow and cannot be scaled to larger datasets, thus we do not recommend this solution. Moreover, 
 ADM-IP gets the best FID and sFID scores, thus, in the rest of this paper, we use the input perturbation approach described in 
Sec.~\ref{subsec.IP alleviate EB} as our basic solution.

\begin{table}[hbt!]
\caption{Comparison of different regularization methods. All the models are tested  using $T = 1,000$ sampling  steps.}
\label{FID Ablation}
\begin{center}
\begin{tabular}{@{}lll@{}}
\toprule
\multirow{2}{*}{Model} & \multicolumn{2}{l}{CIFAR10 32$\times$32} \\ \cmidrule(l){2-3} 
                       & FID            & sFID          \\ \midrule
ADM (baseline)         & 2.99           & 4.76          \\
ADM-GP                 & 2.80           & 4.41          \\
ADM-WD                 & 2.82           & 4.61          \\
ADM-IP                & \textbf{2.76}  & \textbf{4.05} \\ \bottomrule
\end{tabular}
\end{center}
\end{table}

%\subsection{Exposure bias reduction}

Finally, we use ADM-IP to quantify the reduction in the exposure bias following the protocol described in Sec.~\ref{sec.Exposure}. 
 The results reported in Tab.~\ref{stochastic exposure bias} show that ADM-IP leads to a significantly lower exposure bias than ADM, and this difference is larger with  longer  sampling sequences.

\begin{table*}[t]
\caption{Comparison between ADM and ADM-IP using models trained with $T = 1,000$ sampling steps and tested with $T' \leq T$  steps.}
\label{DDPM results}
\begin{center}
% \begin{tabular}{@{}llllllllll@{}}
% \toprule
% \multirow{2}{*}{\begin{tabular}[c]{@{}l@{}}Sampling steps\\ ($T'$)\end{tabular}} & \multirow{2}{*}{Model} & \multicolumn{2}{l}{CIFAR10 32$\times$32} & \multicolumn{2}{l}{ImageNet 32$\times$32} & \multicolumn{2}{l}{LSUN tower 64$\times$64} & \multicolumn{2}{l}{CelebA 64$\times$64} \\ \cmidrule(l){3-10} 
%  &  & FID & sFID & FID & sFID & FID & sFID & FID & sFID \\ \midrule
% \multirow{2}{*}{1,000} & ADM (baseline) & 3.58 & 4.05 & 3.53 & 3.37 & 3.39 & 7.96 & 1.60 & 3.80 \\
%  & ADM-IP (ours)& \textbf{3.25} & \textbf{3.75} & \textbf{2.72} & \textbf{2.44} & \textbf{2.68} & \textbf{6.04} & \textbf{1.31} & \textbf{3.38} \\ \midrule
% \multirow{2}{*}{300} & ADM & 3.47 & 4.12 & 3.52 & 3.67 & 3.13 & 8.39 & 1.82 & 4.25 \\
%  & ADM-IP (ours)& \textbf{3.14} & \textbf{3.72} & \textbf{2.66} & \textbf{2.60} & \textbf{2.60} & \textbf{5.98} & \textbf{1.43} & \textbf{3.36} \\ \midrule
% \multirow{2}{*}{100} & ADM & 3.56 & 4.42 & 4.24 & 4.36 & 3.50 & 11.1 & 3.02 & 5.76 \\
%  & ADM-IP (ours)& \textbf{3.12} & \textbf{3.86} & \textbf{3.22} & \textbf{3.11} & \textbf{2.79} & \textbf{6.56} & \textbf{2.21} & \textbf{4.33} \\ \midrule
% \multirow{2}{*}{80} & ADM & 3.74 & 4.66 & 4.47 & 4.65 & 4.17 & 12.6 & 3.75 & 6.80 \\
%  & ADM-IP (ours)& \textbf{3.26} & \textbf{3.89} & \textbf{3.50} & \textbf{3.36} & \textbf{2.95} & \textbf{6.93} & \textbf{2.67} & \textbf{4.69} \\ \bottomrule
% \end{tabular}
\begin{tabular}{@{}llllllllllll@{}}
\toprule
\multirow{2}{*}{\begin{tabular}[c]{@{}l@{}}Sampling steps\\ ($T'$)\end{tabular}} & \multirow{2}{*}{Model} & \multicolumn{2}{l}{CIFAR10} & \multicolumn{2}{l}{ImageNet 32} & \multicolumn{2}{l}{LSUN tower 64} & \multicolumn{2}{l}{CelebA 64} & \multicolumn{2}{l}{FFHQ 128} \\ \cmidrule(lr){3-4} \cmidrule(lr){5-6} \cmidrule(lr){7-8} \cmidrule(lr){9-10} \cmidrule(lr){11-12}
 &  & FID & sFID & FID & sFID & FID & sFID & FID & sFID & FID & sFID \\ \midrule
\multirow{2}{*}{1,000} & ADM (baseline) & 2.99 & 4.76 & 3.60 & 3.30 & 3.39 & 7.96 & 1.60 & 3.80 & 9.65 & 12.53 \\
 & ADM-IP (ours) & \textbf{2.76} & \textbf{4.05} & \textbf{2.87} & \textbf{2.39} & \textbf{2.68} & \textbf{6.04} & \textbf{1.31} & \textbf{3.38} & \textbf{2.98} & \textbf{5.59} \\ \midrule
\multirow{2}{*}{300} & ADM & 2.95 & 4.95 & 3.58 & 3.48 & 3.31 & 8.39 & 1.82 & 4.25 & 9.55 & 12.60 \\
 & ADM-IP & \textbf{2.67} & \textbf{4.14} & \textbf{2.74} & \textbf{2.58} & \textbf{2.60} & \textbf{5.98} & \textbf{1.43} & \textbf{3.36} & \textbf{3.74} & \textbf{5.97} \\ \midrule
\multirow{2}{*}{100} & ADM & 3.37 & 5.66 & 4.26 & 4.48 & 3.50 & 11.10 & 3.02 & 5.76 & 14.52 & 16.02 \\
 & ADM-IP & \textbf{2.70} & \textbf{4.51} & \textbf{3.24} & \textbf{3.13} & \textbf{2.79} & \textbf{6.56} & \textbf{2.21} & \textbf{4.33} & \textbf{5.94} & \textbf{7.90} \\ \midrule
\multirow{2}{*}{80} & ADM & 3.63 & 5.97 & 4.61 & 4.76 & 4.17 & 12.60 & 3.75 & 6.80 & 17.00 & 18.02 \\
 & ADM-IP & \textbf{2.93} & \textbf{4.69} & \textbf{3.57} & \textbf{3.33} & \textbf{2.95} & \textbf{6.93} & \textbf{2.67} & \textbf{4.69} & \textbf{6.89} & \textbf{8.79} \\ \bottomrule
\end{tabular}
\end{center}
\end{table*}

\subsection{Main results}
\label{sec.main results}

{\bf Comparison with DDPMs.}
We compare ADM-IP with ADM using CIFAR10, ImageNet 32$\times$32, LSUN tower 64$\times$64, CelebA 64$\times$64 \cite{liu2015faceattributes} and FFHQ 128$\times$128. 
Following prior work \citep{DDPMs,IDDPM}, we generate 50K samples for each trained model and we use the full training set to compute the reference distribution statistics, except for LSUN tower where (again following \citep{DDPMs,IDDPM}) we use 50K training samples as the reference data. When training, we always use $T=1,000$ steps for all the models.
At inference time, the results reported with $T' < T$ sampling steps have been obtained using the
 {\em respacing} technique \citep{IDDPM}. 
 As previously mentioned (see Sec.~\ref{subsec.gamma choice}) we keep fixed $\gamma=0.1$ in all the experiments and the datasets.
 We refer to Appendix~\ref{sec:hyperparameters} for
 the complete list of hyperparameters (e.g. the learning rate, the batch size, etc.) 
 and network architecture settings, which are the same for both ADM and ADM-IP. 

The results reported in Tab.~\ref{DDPM results} show that, independently of the dataset and the number of sampling steps ($T' \leq T$), ADM-IP is {\em always}  better than ADM in terms of both the FID and sFID metrics, sometimes drastically better. For instance, on LSUN, with $T'=80$, we have a more than 5 sFID score improvement with respect to ADM. 
On FFHQ 128$\times$128, with $T' = 1,000$, we have almost 7 points of improvement compared to both the FID and the sFID scores.
In addition to the experiments shown in Tab.~\ref{DDPM results}, we used $T'=900$ sampling steps and our ADM-IP 
on CelebA 64$\times$64, achieving a result of  1.27 FID, which is the new state-of-the-art performance for unconditional generation on this dataset.

Note that, for most datasets, both the baseline (ADM) and ADM-IP reach the best results with $T' < T$ (specifically, with $T' \in [100, 300]$). As mentioned in Sec.~\ref{sec.Exposure}, this is most likely a confirmation of the exposure bias problem: a shorter sampling trajectory accumulates a smaller prediction error.

\begin{figure*}[h]
\centering
	\subfigure[ImageNet 32$\times$32]{
		\begin{minipage}{5.8cm}
         \includegraphics[width=\textwidth]{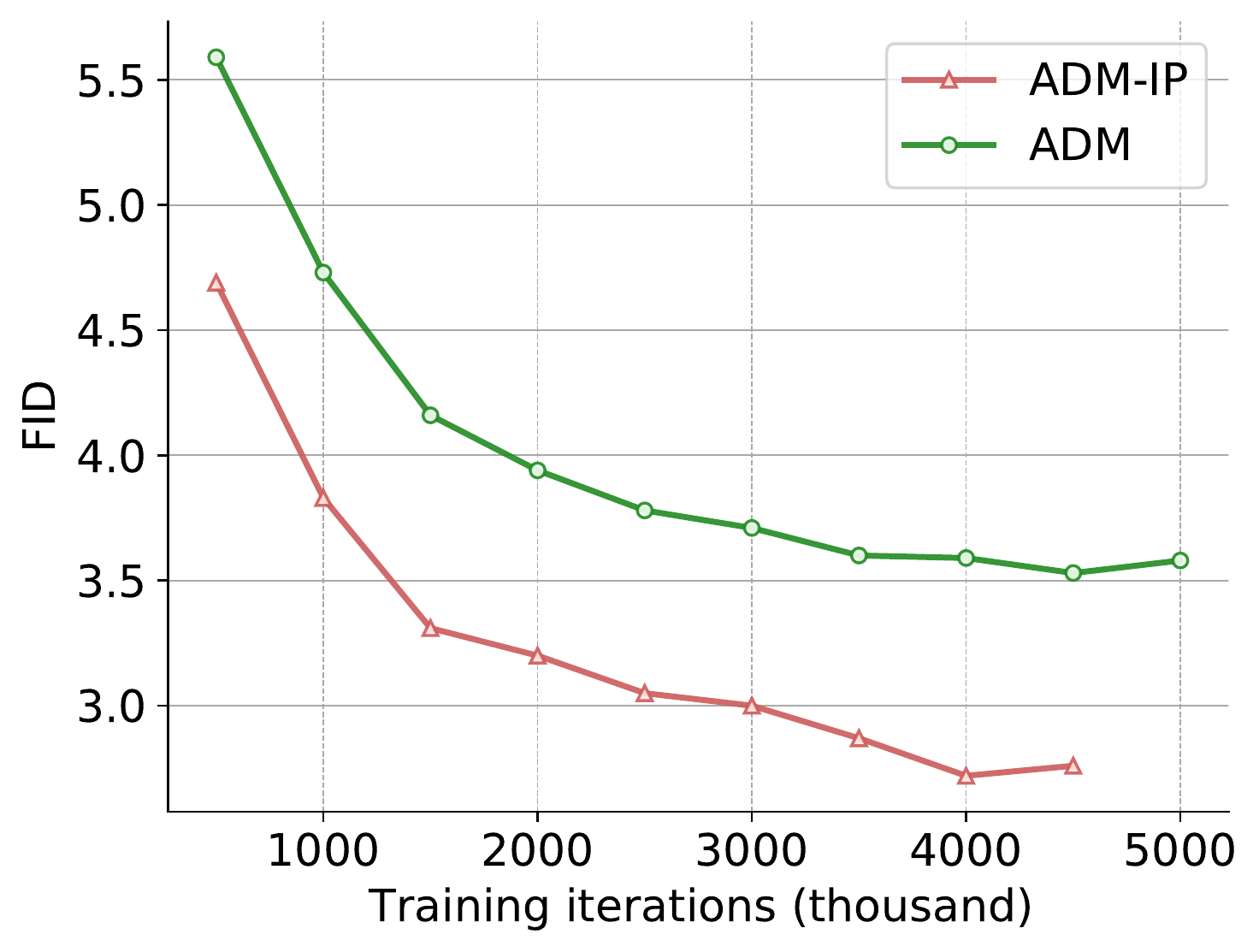} 
         \label{a}
		\end{minipage}
		\hspace{4mm}
	}
	\subfigure[LSUN tower 64$\times$64]{
		\begin{minipage}{5.8cm}
		\includegraphics[width=\textwidth]{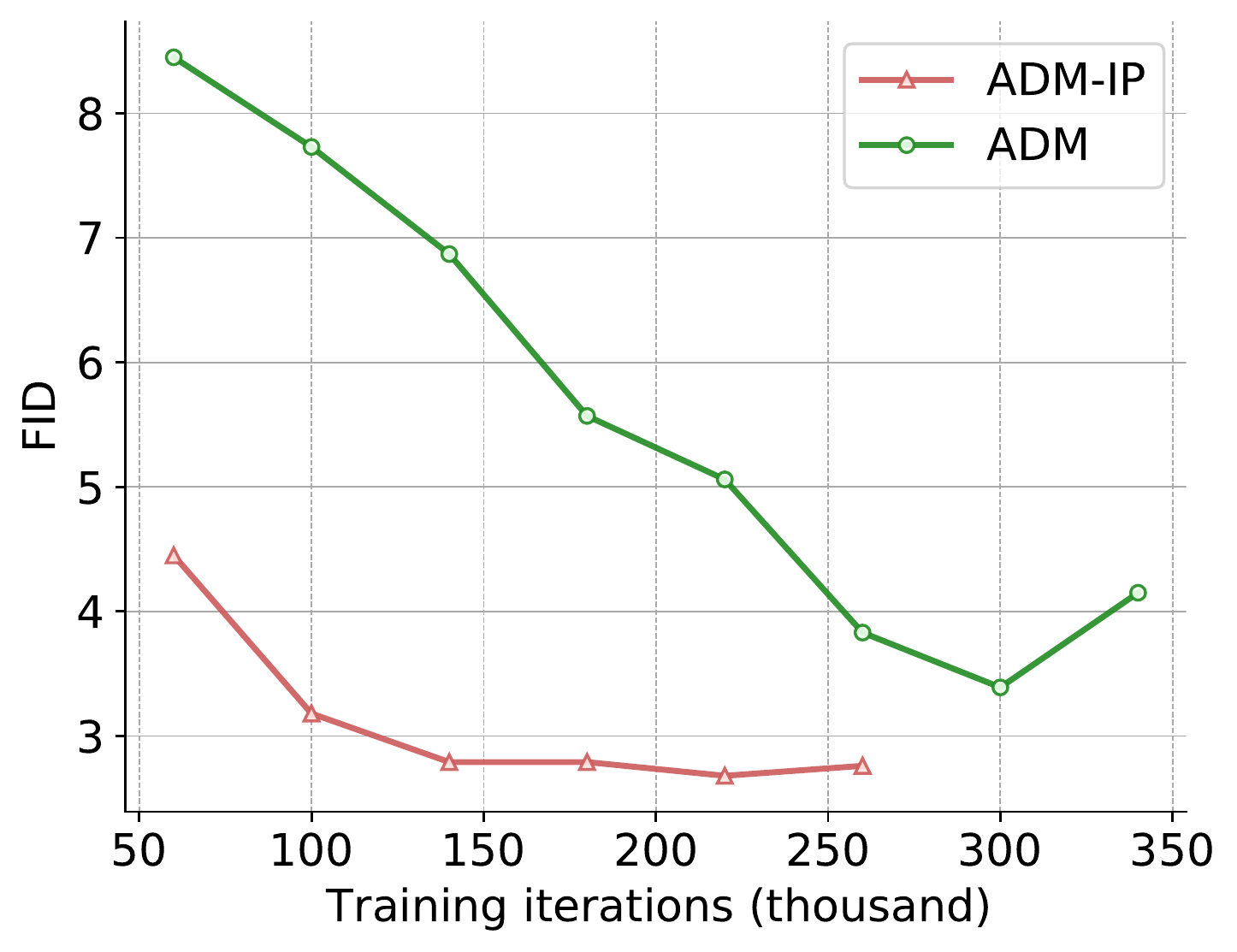}
		\label{b}
		\end{minipage}
		\hspace{4mm}
	}
	\subfigure[CelebA 64$\times$64]{
		\begin{minipage}{5.8cm}
		\includegraphics[width=\textwidth]{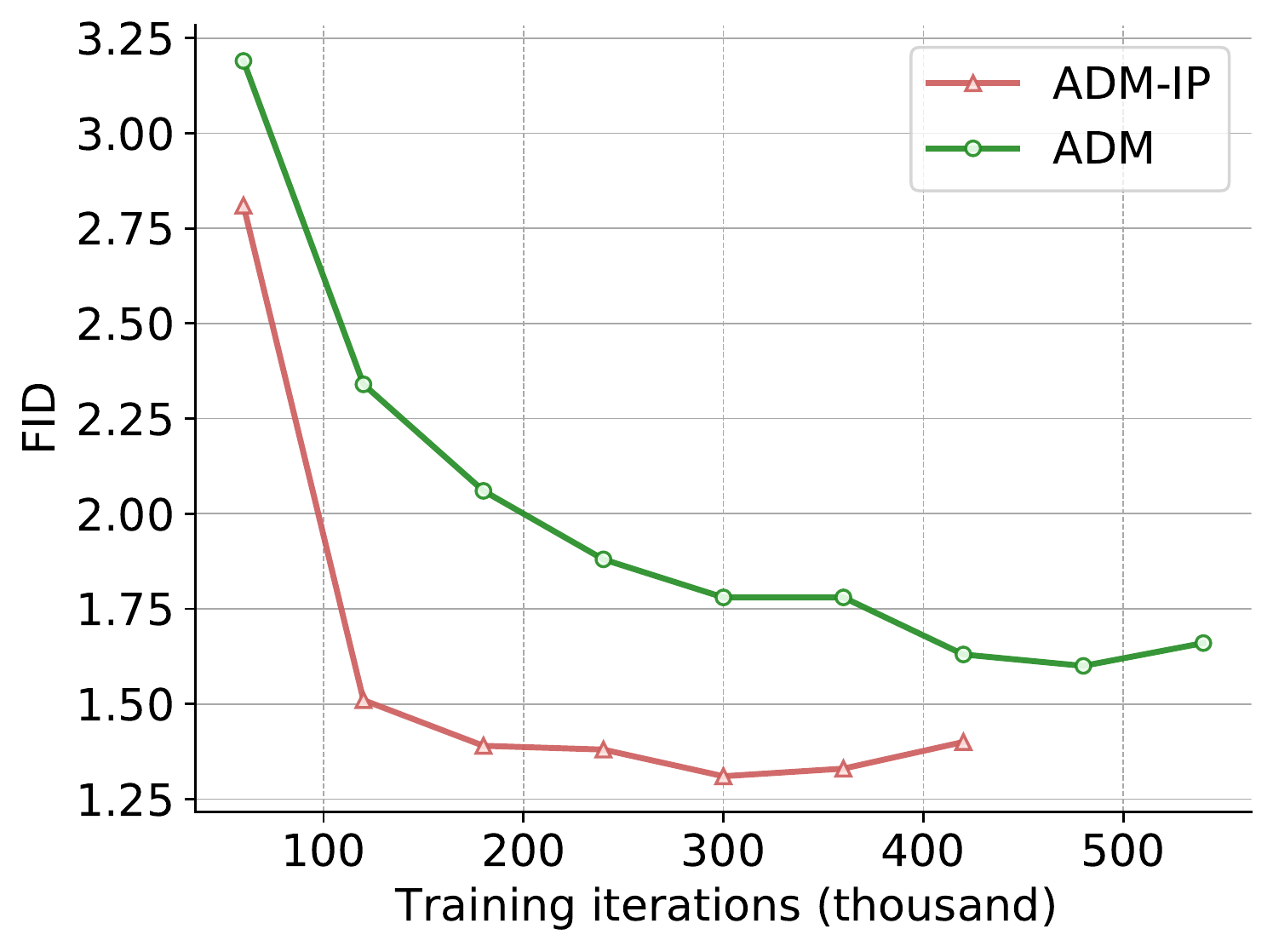}
		\label{c}
		\end{minipage}
		\hspace{4mm}
	}
        \subfigure[FFHQ 128$\times$128]{
		\begin{minipage}{5.8cm}
		\includegraphics[width=\textwidth]{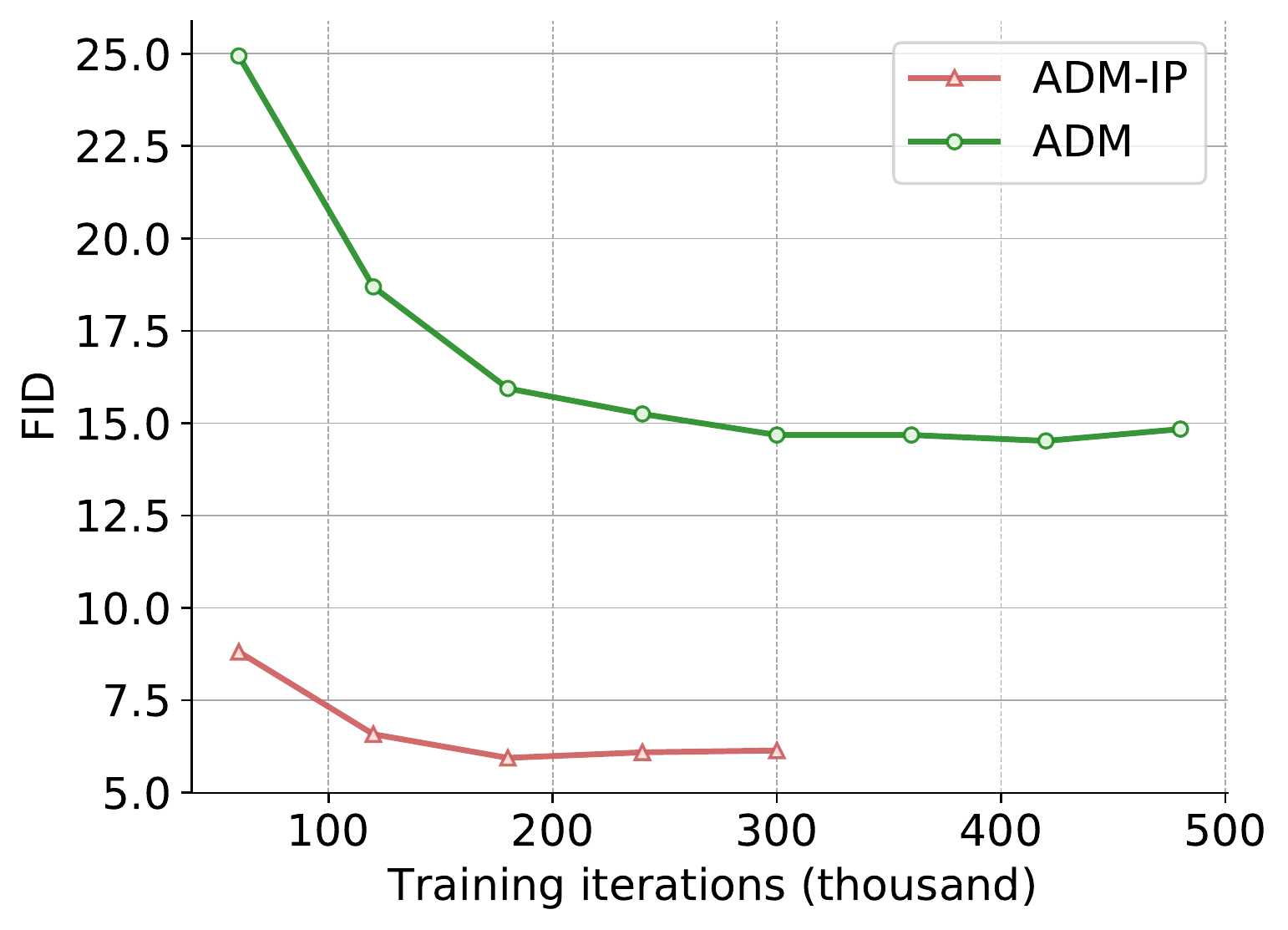}
		\label{d}
		\end{minipage}
		\hspace{4mm}
	}
\caption{FID scores with respect to the number of  training iterations. Each FID value is computed using $T' =1,000$ inference sampling steps, except for the FFHQ dataset, for which we used $T' =100$.
} 
\label{FID curve}
\end{figure*}

\begin{figure}[ht]
\vskip 0.2in
\begin{center}
\centerline{\includegraphics[width=0.8\columnwidth]{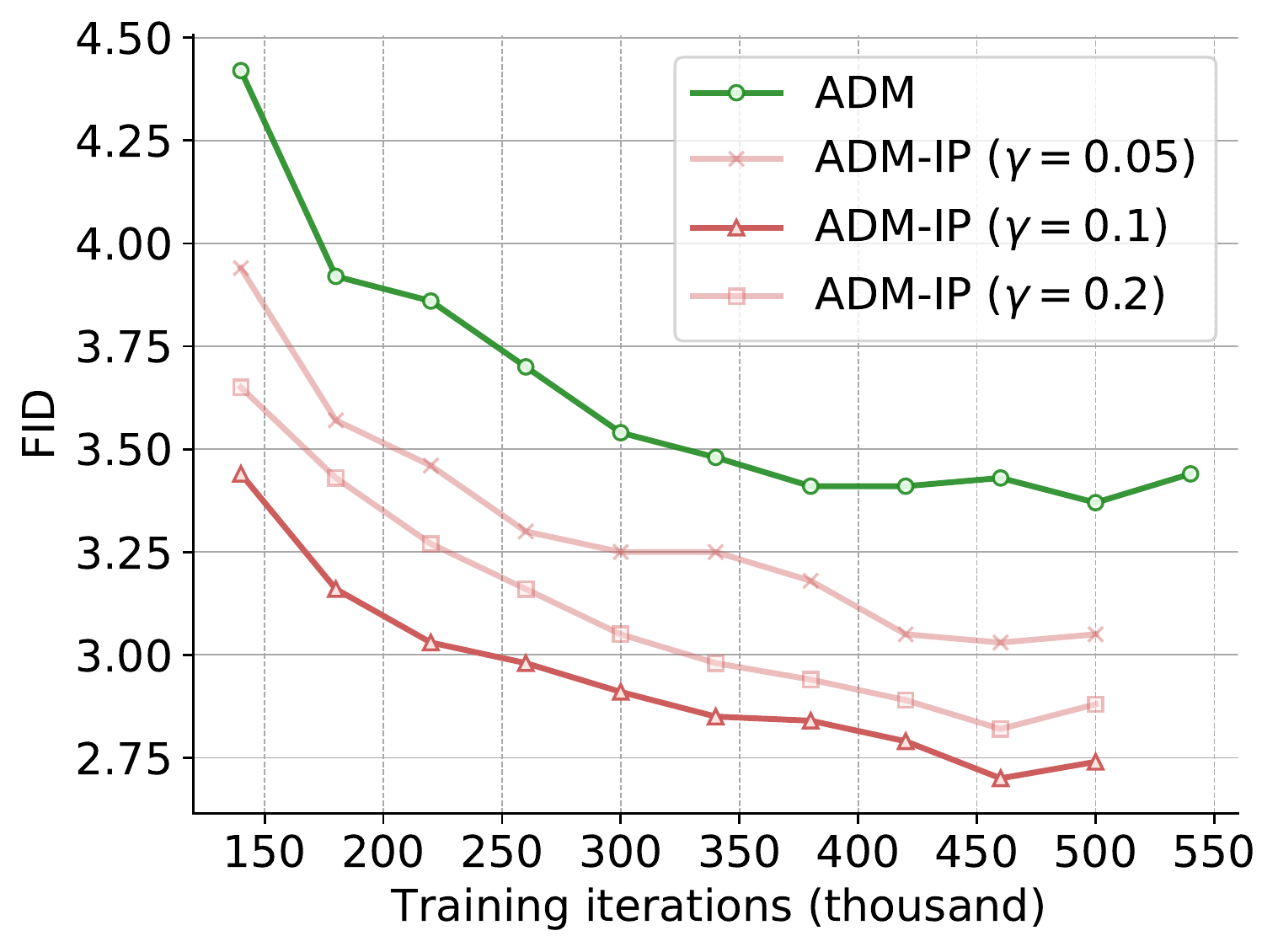}}
\caption{CIFAR10: FID scores with respect to the number of  training iterations with different $\gamma$ values. Each FID score is computed using $T' = 100$ inference sampling steps.
}
\label{cifar_gamma}
\end{center}
\vskip -0.2in
\end{figure}

Besides generating significantly better images, ADM-IP converges much faster than the baseline  during training in
all the five datasets (see Fig.~\ref{FID curve} and \ref{cifar_gamma}).
For instance, on
LSUN tower and CelebA,  ADM-IP converges at 220K and 300K training iterations while ADM saturates around 300K and 480K iterations, respectively. 
Fig.~\ref{FID curve} shows also that,
even before convergence, ADM-IP quickly beats the ADM results obtained when the latter has converged. For instance, on CelebA, 
ADM-IP gets FID 1.51 at 120K training iterations, whereas ADM gets FID 1.6 at convergence (480K iterations), exhibiting a 4x training speed-up. 
On the larger resolution  FFHQ dataset, ADM receives FID 14.52 at convergence (420K iterations), while ADM-IP achieves a FID score of 8.81 with only 60K iterations: an improvement of 5.71 points with a 7x training speed-up.
Fig.~\ref{cifar_gamma} shows a similar trend for the CIFAR10 dataset. In this figure, we also plot the results of ADM-IP with different $\gamma$ values (Sec.~\ref{subsec.gamma choice}).
 
The training iterations until convergence for each model are summarized in Tab.~\ref{acceleration}. The much faster convergence of our method is most likely due to the regularization effect of the input perturbation. 
In fact, as commonly happens with regularization techniques \cite{mixup,OurNIPS,https://doi.org/10.48550/arxiv.2204.03632}, the proposed input perturbation also introduces an inductive bias in training. In our case, it is: close points in the domain of the prediction function should lead to similar outcomes. Our empirical results show that this bias helps the DDPM training.

Tab.~\ref{acceleration} also shows that ADM-IP can drastically accelerate the inference process, i.e. obtaining better results than the baseline with shorter sampling trajectories. For example, with only 60 or 80 steps, ADM-IP gets a better or an equivalent FID than ADM  (tested with the standard  1,000 sampling steps) on all datasets, except for CelebA, where ADM-IP needs 200 sampling steps to reach the same result. This comparison shows a remarkable 5x to 16.7x speed-up of the inference stage, which is particularly significant for the larger resolution  FFHQ dataset. 

Finally, we measure the recall and precision for the generated samples using the method in \citet{kynkaanniemi2019improved}. The results show that the recall and precision achieved by ADM and ADM-IP have no significant difference, which indicates that our input perturbation does not affect the sample diversity (see Appendix~\ref{sec:recall and precision}).

\begin{table}[h]
\caption{ADM-IP training and testing acceleration. Note that, 
for a single   training iteration, ADM and ADM-IP take exactly the same amount of time, and  the same is true for a single sampling step.}
\label{acceleration}
\begin{center}
\begin{tabular}{@{}llllll@{}}
\toprule
Dataset & Model & \begin{tabular}[c]{@{}l@{}}Training \\ iterations\end{tabular} & \begin{tabular}[c]{@{}l@{}}Sampling \\ steps\end{tabular} & FID \\ \midrule
\multirow{2}{*}{\begin{tabular}[c]{@{}l@{}}CIFAR10\\ 32$\times$32\end{tabular}} & ADM & 500K & 1,000 & 2.99 \\
 & ADM-IP & 460K & 80 & 2.93 \\ \midrule
\multirow{2}{*}{\begin{tabular}[c]{@{}l@{}}ImageNet\\ 32$\times$32\end{tabular}} & ADM & 4500K & 1,000 & 3.53 \\
 & ADM-IP & 4000K & 80 & 3.50 \\ \midrule
\multirow{2}{*}{\begin{tabular}[c]{@{}l@{}}LSUN tower\\ 64$\times$64\end{tabular}} & ADM & 300K & 1,000 & 3.39 \\
 & ADM-IP & 220K & 60 & 3.31 \\ \midrule
\multirow{2}{*}{\begin{tabular}[c]{@{}l@{}}CelabA\\ 64$\times$64\end{tabular}} & ADM & 480K & 1,000 & 1.60 \\
 & ADM-IP & 300K & 200 & 1.53 \\ \midrule
\multirow{2}{*}{\begin{tabular}[c]{@{}l@{}}FFHQ\\ 128$\times$128\end{tabular}} & ADM & 420K & 1,000 & 9.65 \\
& ADM-IP & 180K & 60 & 8.72 \\ \bottomrule
\end{tabular}
\end{center}
\end{table}

% \begin{figure*}[h]
% \centering
% 	\subfigure[CIFAR10 32$\times$32]{
% 		\begin{minipage}{5.5cm}
%          \includegraphics[width=\textwidth]{images/FID_cifar10_results.png} 
%          \label{a}
% 		\end{minipage}
% 		\hspace{4mm}
% 	}
% 	\subfigure[ImageNet 32$\times$32]{
% 		\begin{minipage}{5.5cm}
% 		\includegraphics[width=\textwidth]{images/FID_imagenet32_results.png}
% 		\label{b}
% 		\end{minipage}
% 		\hspace{4mm}
% 	}
% 	\subfigure[LSUN tower 64$\times$64]{
% 		\begin{minipage}{5.5cm}
% 		\includegraphics[width=\textwidth]{images/FID_LSUN_results.png}
% 		\label{c}
% 		\end{minipage}
% 		\hspace{4mm}
% 	}
%         \subfigure[CelebA 64$\times$64]{
% 		\begin{minipage}{5.5cm}
% 		\includegraphics[width=\textwidth]{images/FID_celeba_results.png}
% 		\label{d}
% 		\end{minipage}
% 		\hspace{4mm}
% 	}
% \caption{FID scores with respect to the number of  training iterations. Each FID value is computed using $T =1,000$ sampling steps.
% } 
% \label{FID curve}
% \end{figure*}

{\bf Comparison with DDIMs.}
In order to show the generality of our proposal, we
use  Alg.~\ref{alg:4}
with the Denoising Diffusion
Implicit Models (DDIMs)
proposed by \citet{DBLP:conf/iclr/SongME21} (Sec.~\ref{sec.Related}).  We train both the baseline (DDIM) and our method (DDIM-IP) on CIFAR10
using the public code provided by \citet{DBLP:conf/iclr/SongME21}. 
Since  training with DDIM is particularly slow, we use only CIFAR10 for this comparison.
We use the default hyperparameters settings (e.g. $T=1,000$) in their code and train both models for 1,600K iterations with batch size  128. We test the performance of the two models with both $\eta=0$ and $\eta=0.5$, where $\eta$ is the coefficient of stochasticity sampling  in DDIMs.
Also in this case, for our method (DDIM-IP) we use $\gamma = 0.1$ {\em without any fine-tuning}.

We report the  results in Tab.~\ref{DDIM results}, which show that DDIM-IP consistently obtains  better FID scores than DDIM in {\em all} conditions (i.e., independently of the number of sampling steps and the value of  $\eta$). Importantly, the fewer the sampling steps, the more the FID gain which is obtained with  input perturbation. For instance, with $\eta=0.5$, the FID gain of DDIM-IP  is 7.16 with 10 sampling steps versus 0.89 with 1,000 sampling steps. 
Analogously,  with $\eta=0$ and 10 sampling steps,  DDIM-IP drastically improves DDIM with a 3.67 FID margin.
Since the main advantage of DDIMs with respect to DDPMs is their reduced number of  sampling steps \cite{DBLP:conf/iclr/SongME21}, and they indeed are mainly used for accelerating the inference stage, input perturbation greatly matches this goal, and it significantly improves the sample quality of the implicit models in a short  sampling sequence regime.

\begin{table}[h]
\caption{CIFAR10: Comparison between DDIM and DDIM-IP using models trained with $T = 1,000$ sampling steps and tested with $T' \leq T$  steps.}
\label{DDIM results}
\begin{center}
\begin{tabular}{@{}lllllll@{}}
\toprule
\multirow{2}{*}{$\eta$} & \multirow{2}{*}{Model} & \multicolumn{5}{c}{Sampling steps ($T'$)} \\ \cmidrule(l){3-7} 
 &  & 10 & 20 & 50 & 100 & 1,000 \\ \midrule
\multirow{2}{*}{0} & DDIM & 14.21 & 7.50 & 5.17 & 4.66 & 4.29 \\
 & DDIM-IP & \textbf{10.54} & \textbf{5.70} & \textbf{4.66} & \textbf{4.52} & \textbf{4.27} \\ \midrule
\multirow{2}{*}{0.5} & DDIM & 17.24 & 8.87 & 5.59 & 4.88 & 4.45 \\
 & DDIM-IP & \textbf{10.06} & \textbf{5.53} & \textbf{3.95} & \textbf{3.66} & \textbf{3.56} \\ \bottomrule
\end{tabular}
\end{center}
\end{table}

\section{Conclusions}
\label{sec.Conclusions}

In this paper, we proposed DDPM-IP, a  regularization method  for DDPM training which is based on input perturbation to explicitly model the prediction errors and alleviate the DDPM exposure bias problem. We empirically showed that DDPM-IP can significantly improve image quality and drastically reduce both the training and the inference time. The proposed method is straightforward and does not require any change in the network architecture or the specific loss function. This simplicity makes it very easy to be reproduced and plugged into existing DDPMs. Although we tested DDPM-IP only on an image domain, there are no domain-specific assumptions behind our method, hence we presume it can be more generally applied to other domains. 

{\bf Limitations.} Since training DDPMs is very computationally heavy, in this paper we used only datasets with small resolution images. 
We leave the extension of our experiments to larger resolution images (and corresponding larger backbone networks) as a future work.
However, we emphasize that our best results have been obtained with FFHQ 128$\times$128, which is the dataset with the largest resolution images we tested, which probably confirms that our regularization method is specifically effective with higher dimensional input spaces.

\section*{Acknowledgments}
This work has been supported by the European Union’s Horizon 2020 research and innovation programme under the Marie Skłodowska-Curie grant agreement No 955778.
Moreover, we acknowledge the CINECA award under the ISCRA initiative, for the availability of high-performance computing resources and support.


\bibliography{example_paper}

\begin{thebibliography}{57}
\providecommand{\natexlab}[1]{#1}
\providecommand{\url}[1]{\texttt{#1}}
\expandafter\ifx\csname urlstyle\endcsname\relax
  \providecommand{\doi}[1]{doi: #1}\else
  \providecommand{\doi}{doi: \begingroup \urlstyle{rm}\Url}\fi

\bibitem[Balestriero et~al.(2022)Balestriero, Bottou, and
  LeCun]{https://doi.org/10.48550/arxiv.2204.03632}
Balestriero, R., Bottou, L., and LeCun, Y.
\newblock The effects of regularization and data augmentation are class
  dependent.
\newblock \emph{arXiv:2204.03632}, 2022.

\bibitem[Bengio et~al.(2015)Bengio, Vinyals, Jaitly, and
  Shazeer]{bengio2015scheduled}
Bengio, S., Vinyals, O., Jaitly, N., and Shazeer, N.
\newblock {Scheduled sampling for sequence prediction with recurrent neural
  networks}.
\newblock In \emph{NeurIPS}, 2015.

\bibitem[Brock et~al.(2018)Brock, Donahue, and Simonyan]{BigGAN}
Brock, A., Donahue, J., and Simonyan, K.
\newblock Large scale gan training for high fidelity natural image synthesis.
\newblock \emph{arXiv preprint arXiv:1809.11096}, 2018.

\bibitem[Chapelle et~al.(2000)Chapelle, Weston, Bottou, and
  Vapnik]{DBLP:conf/nips/ChapelleWBV00}
Chapelle, O., Weston, J., Bottou, L., and Vapnik, V.
\newblock Vicinal risk minimization.
\newblock In \emph{NIPS}, 2000.

\bibitem[Chen et~al.(2021)Chen, Zhang, Zen, Weiss, Norouzi, and
  Chan]{DBLP:conf/iclr/ChenZZWNC21}
Chen, N., Zhang, Y., Zen, H., Weiss, R.~J., Norouzi, M., and Chan, W.
\newblock Wavegrad: Estimating gradients for waveform generation.
\newblock In \emph{ICLR}, 2021.

\bibitem[Chrabaszcz et~al.(2017)Chrabaszcz, Loshchilov, and Hutter]{imagenet32}
Chrabaszcz, P., Loshchilov, I., and Hutter, F.
\newblock A downsampled variant of imagenet as an alternative to the cifar
  datasets.
\newblock \emph{arXiv:1707.08819}, 2017.

\bibitem[Croitoru et~al.(2022)Croitoru, Hondru, Ionescu, and Shah]{survey2}
Croitoru, F.-A., Hondru, V., Ionescu, R.~T., and Shah, M.
\newblock Diffusion models in vision: A survey.
\newblock \emph{arXiv preprint arXiv:2209.04747}, 2022.

\bibitem[Dhariwal \& Nichol(2021)Dhariwal and Nichol]{ADM}
Dhariwal, P. and Nichol, A.~Q.
\newblock Diffusion models beat {GANs} on image synthesis.
\newblock In \emph{NeurIPS}, 2021.

\bibitem[Goodfellow et~al.(2014)Goodfellow, Pouget-Abadie, Mirza, Xu,
  Warde-Farley, Ozair, Courville, and Bengio]{goodfellow2014generative}
Goodfellow, I., Pouget-Abadie, J., Mirza, M., Xu, B., Warde-Farley, D., Ozair,
  S., Courville, A., and Bengio, Y.
\newblock Generative adversarial nets.
\newblock In \emph{NeurIPS}, 2014.

\bibitem[Goodfellow et~al.(2016)Goodfellow, Bengio, and
  Courville]{DeepLearning}
Goodfellow, I., Bengio, Y., and Courville, A.
\newblock \emph{Deep Learning}.
\newblock MIT Press, 2016.
\newblock \url{http://www.deeplearningbook.org}.

\bibitem[Gu et~al.(2021)Gu, Chen, Bao, Wen, Zhang, Chen, Yuan, and
  Guo]{DBLP:journals/corr/abs-2111-14822}
Gu, S., Chen, D., Bao, J., Wen, F., Zhang, B., Chen, D., Yuan, L., and Guo, B.
\newblock Vector quantized diffusion model for text-to-image synthesis.
\newblock \emph{arXiv:2111.14822}, 2021.

\bibitem[Gulrajani et~al.(2017)Gulrajani, Ahmed, Arjovsky, Dumoulin, and
  Courville]{WGAN-GP}
Gulrajani, I., Ahmed, F., Arjovsky, M., Dumoulin, V., and Courville, A.~C.
\newblock Improved training of wasserstein gans.
\newblock \emph{NeurIPS}, 30, 2017.

\bibitem[Heusel et~al.(2017)Heusel, Ramsauer, Unterthiner, Nessler, and
  Hochreiter]{FID}
Heusel, M., Ramsauer, H., Unterthiner, T., Nessler, B., and Hochreiter, S.
\newblock Gans trained by a two time-scale update rule converge to a local nash
  equilibrium.
\newblock \emph{NeurIPS}, 30, 2017.

\bibitem[Ho et~al.(2020)Ho, Jain, and Abbeel]{DDPMs}
Ho, J., Jain, A., and Abbeel, P.
\newblock Denoising diffusion probabilistic models.
\newblock In \emph{NeurIPS}, 2020.

\bibitem[Hoogeboom et~al.(2021)Hoogeboom, Nielsen, Jaini, Forr{\'{e}}, and
  Welling]{DBLP:conf/nips/HoogeboomNJFW21}
Hoogeboom, E., Nielsen, D., Jaini, P., Forr{\'{e}}, P., and Welling, M.
\newblock Argmax flows and multinomial diffusion: Learning categorical
  distributions.
\newblock In \emph{NeurIPS}, 2021.

\bibitem[Hoogeboom et~al.(2022)Hoogeboom, Gritsenko, Bastings, Poole, van~den
  Berg, and Salimans]{DBLP:conf/iclr/HoogeboomGBPBS22}
Hoogeboom, E., Gritsenko, A.~A., Bastings, J., Poole, B., van~den Berg, R., and
  Salimans, T.
\newblock Autoregressive diffusion models.
\newblock In \emph{ICLR}, 2022.

\bibitem[Karras et~al.(2019)Karras, Laine, and Aila]{karras2019style}
Karras, T., Laine, S., and Aila, T.
\newblock A style-based generator architecture for generative adversarial
  networks.
\newblock In \emph{Proceedings of the IEEE/CVF conference on computer vision
  and pattern recognition}, pp.\  4401--4410, 2019.

\bibitem[Karras et~al.(2022)Karras, Aittala, Aila, and
  Laine]{karras2022elucidating}
Karras, T., Aittala, M., Aila, T., and Laine, S.
\newblock Elucidating the design space of diffusion-based generative models.
\newblock In \emph{NeurIPS}, 2022.

\bibitem[Kong \& Ping(2021)Kong and Ping]{kong2021fast}
Kong, Z. and Ping, W.
\newblock On fast sampling of diffusion probabilistic models.
\newblock \emph{arXiv preprint arXiv:2106.00132}, 2021.

\bibitem[Krizhevsky et~al.(2009)Krizhevsky, Hinton, et~al.]{cifar10}
Krizhevsky, A., Hinton, G., et~al.
\newblock Learning multiple layers of features from tiny images.
\newblock 2009.

\bibitem[Krogh \& Hertz(1991)Krogh and Hertz]{WeightDecay}
Krogh, A. and Hertz, J.
\newblock A simple weight decay can improve generalization.
\newblock \emph{NeurIPS}, 4, 1991.

\bibitem[Kynk{\"a}{\"a}nniemi et~al.(2019)Kynk{\"a}{\"a}nniemi, Karras, Laine,
  Lehtinen, and Aila]{kynkaanniemi2019improved}
Kynk{\"a}{\"a}nniemi, T., Karras, T., Laine, S., Lehtinen, J., and Aila, T.
\newblock Improved precision and recall metric for assessing generative models.
\newblock \emph{Advances in Neural Information Processing Systems}, 32, 2019.

\bibitem[Liu et~al.(2022)Liu, Williams, Jacobson, Fidler, and
  Litany]{LipsWeightDecay}
Liu, H.-T.~D., Williams, F., Jacobson, A., Fidler, S., and Litany, O.
\newblock Learning smooth neural functions via lipschitz regularization.
\newblock \emph{arXiv preprint arXiv:2202.08345}, 2022.

\bibitem[Liu et~al.(2021)Liu, Sangineto, Bi, Sebe, Lepri, and Nadai]{OurNIPS}
Liu, Y., Sangineto, E., Bi, W., Sebe, N., Lepri, B., and Nadai, M.~D.
\newblock Efficient training of visual transformers with small datasets.
\newblock \emph{NeurIPS}, 2021.

\bibitem[Liu et~al.(2015)Liu, Luo, Wang, and Tang]{liu2015faceattributes}
Liu, Z., Luo, P., Wang, X., and Tang, X.
\newblock Deep learning face attributes in the wild.
\newblock In \emph{Proceedings of International Conference on Computer Vision
  (ICCV)}, December 2015.

\bibitem[Loshchilov \& Hutter(2019)Loshchilov and Hutter]{AdamW}
Loshchilov, I. and Hutter, F.
\newblock Decoupled weight decay regularization.
\newblock \emph{ICLR}, 2019.

\bibitem[Luo(2022)]{UnderstandingDM}
Luo, C.
\newblock Understanding diffusion models: A unified perspective.
\newblock \emph{arXiv preprint arXiv:2208.11970}, 2022.

\bibitem[Micikevicius et~al.(2017)Micikevicius, Narang, Alben, Diamos, Elsen,
  Garcia, Ginsburg, Houston, Kuchaiev, Venkatesh, et~al.]{loss_scaling}
Micikevicius, P., Narang, S., Alben, J., Diamos, G., Elsen, E., Garcia, D.,
  Ginsburg, B., Houston, M., Kuchaiev, O., Venkatesh, G., et~al.
\newblock Mixed precision training.
\newblock \emph{arXiv:1710.03740}, 2017.

\bibitem[Mittal et~al.(2021)Mittal, Engel, Hawthorne, and
  Simon]{DBLP:conf/ismir/MittalEHS21}
Mittal, G., Engel, J.~H., Hawthorne, C., and Simon, I.
\newblock Symbolic music generation with diffusion models.
\newblock In \emph{Proceedings of the 22nd International Society for Music
  Information Retrieval Conference, {ISMIR}}, 2021.

\bibitem[Miyato et~al.(2018)Miyato, Kataoka, Koyama, and
  Yoshida]{SpectralNorm2018}
Miyato, T., Kataoka, T., Koyama, M., and Yoshida, Y.
\newblock Spectral normalization for generative adversarial networks.
\newblock \emph{arXiv preprint arXiv:1802.05957}, 2018.

\bibitem[Nash et~al.(2021)Nash, Menick, Dieleman, and Battaglia]{sFID}
Nash, C., Menick, J., Dieleman, S., and Battaglia, P.~W.
\newblock Generating images with sparse representations.
\newblock \emph{arXiv:2103.03841}, 2021.

\bibitem[Nichol \& Dhariwal(2021)Nichol and Dhariwal]{IDDPM}
Nichol, A.~Q. and Dhariwal, P.
\newblock Improved denoising diffusion probabilistic models.
\newblock In \emph{ICML}, 2021.

\bibitem[Nichol et~al.(2022)Nichol, Dhariwal, Ramesh, Shyam, Mishkin, McGrew,
  Sutskever, and Chen]{DBLP:conf/icml/NicholDRSMMSC22}
Nichol, A.~Q., Dhariwal, P., Ramesh, A., Shyam, P., Mishkin, P., McGrew, B.,
  Sutskever, I., and Chen, M.
\newblock {GLIDE:} towards photorealistic image generation and editing with
  text-guided diffusion models.
\newblock In \emph{ICML}, 2022.

\bibitem[Paszke et~al.(2019)Paszke, Gross, Massa, Lerer, Bradbury, Chanan,
  Killeen, Lin, Gimelshein, Antiga, et~al.]{pytorch}
Paszke, A., Gross, S., Massa, F., Lerer, A., Bradbury, J., Chanan, G., Killeen,
  T., Lin, Z., Gimelshein, N., Antiga, L., et~al.
\newblock Pytorch: An imperative style, high-performance deep learning library.
\newblock \emph{NeurIPS}, 32, 2019.

\bibitem[Radford et~al.(2021)Radford, Kim, Hallacy, Ramesh, Goh, Agarwal,
  Sastry, Askell, Mishkin, Clark, Krueger, and Sutskever]{radford2021learning}
Radford, A., Kim, J.~W., Hallacy, C., Ramesh, A., Goh, G., Agarwal, S., Sastry,
  G., Askell, A., Mishkin, P., Clark, J., Krueger, G., and Sutskever, I.
\newblock {Learning Transferable Visual Models From Natural Language
  Supervision}.
\newblock In \emph{ICML}, 2021.

\bibitem[Ramesh et~al.(2022)Ramesh, Dhariwal, Nichol, Chu, and Chen]{DALL-E-2}
Ramesh, A., Dhariwal, P., Nichol, A., Chu, C., and Chen, M.
\newblock Hierarchical text-conditional image generation with {CLIP} latents.
\newblock \emph{arXiv:2204.06125}, 2022.

\bibitem[Ranzato et~al.(2016)Ranzato, Chopra, Auli, and
  Zaremba]{DBLP:journals/corr/RanzatoCAZ15}
Ranzato, M., Chopra, S., Auli, M., and Zaremba, W.
\newblock Sequence level training with recurrent neural networks.
\newblock In \emph{ICLR}, 2016.

\bibitem[Rasul et~al.(2021)Rasul, Seward, Schuster, and
  Vollgraf]{rasul2021autoregressive}
Rasul, K., Seward, C., Schuster, I., and Vollgraf, R.
\newblock Autoregressive denoising diffusion models for multivariate
  probabilistic time series forecasting.
\newblock In \emph{ICML}, 2021.

\bibitem[Rennie et~al.(2017)Rennie, Marcheret, Mroueh, Ross, and
  Goel]{rennie2017self}
Rennie, S.~J., Marcheret, E., Mroueh, Y., Ross, J., and Goel, V.
\newblock Self-critical sequence training for image captioning.
\newblock In \emph{CVPR}, 2017.

\bibitem[Rifai et~al.(2011)Rifai, Pascal~Vincent, Glorot, and
  Bengio]{ContractiveAutoencoder}
Rifai, S., Pascal~Vincent, X.~M., Glorot, X., and Bengio, Y.
\newblock Contractive auto-encoders: Explicit invariance during feature
  extraction.
\newblock In \emph{ICML}, 2011.

\bibitem[Rombach et~al.(2021)Rombach, Blattmann, Lorenz, Esser, and
  Ommer]{stableDiffusion}
Rombach, R., Blattmann, A., Lorenz, D., Esser, P., and Ommer, B.
\newblock High-resolution image synthesis with latent diffusion models, 2021.

\bibitem[Saharia et~al.(2022)Saharia, Chan, Saxena, Li, Whang, Denton,
  Ghasemipour, Ayan, Mahdavi, Lopes, et~al.]{Imagen}
Saharia, C., Chan, W., Saxena, S., Li, L., Whang, J., Denton, E., Ghasemipour,
  S. K.~S., Ayan, B.~K., Mahdavi, S.~S., Lopes, R.~G., et~al.
\newblock Photorealistic text-to-image diffusion models with deep language
  understanding.
\newblock \emph{arXiv preprint arXiv:2205.11487}, 2022.

\bibitem[Salimans \& Ho(2022)Salimans and Ho]{DBLP:conf/iclr/SalimansH22}
Salimans, T. and Ho, J.
\newblock Progressive distillation for fast sampling of diffusion models.
\newblock In \emph{ICLR}, 2022.

\bibitem[Schmidt(2019)]{DBLP:conf/emnlp/Schmidt19}
Schmidt, F.
\newblock Generalization in generation: {A} closer look at exposure bias.
\newblock In \emph{Proceedings of the 3rd Workshop on Neural Generation and
  Translation@EMNLP-IJCNLP}, 2019.

\bibitem[Shapiro \& Wilk(1965)Shapiro and Wilk]{shapiro1965analysis}
Shapiro, S.~S. and Wilk, M.~B.
\newblock An analysis of variance test for normality (complete samples).
\newblock \emph{Biometrika}, 52\penalty0 (3/4):\penalty0 591--611, 1965.

\bibitem[Sohl-Dickstein et~al.(2015)Sohl-Dickstein, Weiss, Maheswaranathan, and
  Ganguli]{pmlr-v37-sohl-dickstein15}
Sohl-Dickstein, J., Weiss, E., Maheswaranathan, N., and Ganguli, S.
\newblock Deep unsupervised learning using nonequilibrium thermodynamics.
\newblock In \emph{ICML}, 2015.

\bibitem[Song et~al.(2021{\natexlab{a}})Song, Meng, and
  Ermon]{DBLP:conf/iclr/SongME21}
Song, J., Meng, C., and Ermon, S.
\newblock Denoising diffusion implicit models.
\newblock In \emph{ICLR}, 2021{\natexlab{a}}.

\bibitem[Song \& Ermon(2019)Song and Ermon]{NCSN}
Song, Y. and Ermon, S.
\newblock Generative modeling by estimating gradients of the data distribution.
\newblock \emph{NeurIPS}, 32, 2019.

\bibitem[Song et~al.(2021{\natexlab{b}})Song, Sohl-Dickstein, Kingma, Kumar,
  Ermon, and Poole]{VPVE}
Song, Y., Sohl-Dickstein, J., Kingma, D.~P., Kumar, A., Ermon, S., and Poole,
  B.
\newblock Score-based generative modeling through stochastic differential
  equations.
\newblock In \emph{ICLR}, 2021{\natexlab{b}}.

\bibitem[Welling \& Teh(2011)Welling and Teh]{DBLP:conf/icml/WellingT11}
Welling, M. and Teh, Y.~W.
\newblock Bayesian learning via stochastic gradient langevin dynamics.
\newblock In \emph{ICML}, 2011.

\bibitem[Xiao et~al.(2022)Xiao, Kreis, and Vahdat]{xiao2021tackling}
Xiao, Z., Kreis, K., and Vahdat, A.
\newblock Tackling the generative learning trilemma with denoising diffusion
  {GAN}s.
\newblock In \emph{International Conference on Learning Representations
  (ICLR)}, 2022.

\bibitem[Xu et~al.(2022)Xu, Liu, Tegmark, and Jaakkola]{xu2022poisson}
Xu, Y., Liu, Z., Tegmark, M., and Jaakkola, T.
\newblock Poisson flow generative models.
\newblock \emph{arXiv preprint arXiv:2209.11178}, 2022.

\bibitem[Yang et~al.(2022)Yang, Zhang, and Hong]{survey}
Yang, L., Zhang, Z., and Hong, S.
\newblock Diffusion models: A comprehensive survey of methods and applications.
\newblock \emph{arXiv:2209.00796}, 2022.

\bibitem[Yang et~al.(2019)Yang, Dai, Yang, Carbonell, Salakhutdinov, and
  Le]{DBLP:conf/nips/YangDYCSL19}
Yang, Z., Dai, Z., Yang, Y., Carbonell, J.~G., Salakhutdinov, R., and Le, Q.~V.
\newblock Xlnet: Generalized autoregressive pretraining for language
  understanding.
\newblock In \emph{NeurIPS}, 2019.

\bibitem[Yoshida \& Miyato(2017)Yoshida and Miyato]{SpectralNorm2017}
Yoshida, Y. and Miyato, T.
\newblock Spectral norm regularization for improving the generalizability of
  deep learning.
\newblock \emph{arXiv preprint arXiv:1705.10941}, 2017.

\bibitem[Yu et~al.(2015)Yu, Seff, Zhang, Song, Funkhouser, and Xiao]{LSUN}
Yu, F., Seff, A., Zhang, Y., Song, S., Funkhouser, T., and Xiao, J.
\newblock Lsun: Construction of a large-scale image dataset using deep learning
  with humans in the loop.
\newblock \emph{arXiv preprint arXiv:1506.03365}, 2015.

\bibitem[Zhang et~al.(2018)Zhang, Ciss{\'{e}}, Dauphin, and Lopez{-}Paz]{mixup}
Zhang, H., Ciss{\'{e}}, M., Dauphin, Y.~N., and Lopez{-}Paz, D.
\newblock mixup: Beyond empirical risk minimization.
\newblock In \emph{ICLR}, 2018.

\end{thebibliography}
\bibliographystyle{icml2023}

%%%%%%%%%%%%%%%%%%%%%%%%%%%%%%%%%%%%%%%%%%%%%%%%%%%%%%%%%%%%%%%%%%%%%%%%%%%%%%%
%%%%%%%%%%%%%%%%%%%%%%%%%%%%%%%%%%%%%%%%%%%%%%%%%%%%%%%%%%%%%%%%%%%%%%%%%%%%%%%
% APPENDIX
%%%%%%%%%%%%%%%%%%%%%%%%%%%%%%%%%%%%%%%%%%%%%%%%%%%%%%%%%%%%%%%%%%%%%%%%%%%%%%%
%%%%%%%%%%%%%%%%%%%%%%%%%%%%%%%%%%%%%%%%%%%%%%%%%%%%%%%%%%%%%%%%%%%%%%%%%%%%%%%
\newpage
\appendix
\onecolumn
\section{Appendix}

\subsection{Exposure Bias Analysis}
\label{apendix-exposure-bias}

In this section, we repeat the experiment in Sec.~\ref{sec.Exposure} by removing the randomness component of the sampling process in order to isolate the error of the reverse process which is due only to the prediction network. Specifically, 
we use again ADM \citep{ADM} (trained  with $T=1,000$)  and   ImageNet 32$\times$32,
and we directly measure the difference between a ground truth real image $\pmb{x}_{0}$ and the predicted $\hat{\pmb{x}}_{0}$
using a {\em deterministic sampling}, described in 
 Alg.~\ref{alg:3}. In more detail, given a real image $\pmb{x}_{0}$, we first compute $\pmb{x}_{t}$ by Eq.~\ref{eq.x-t}, then we use the {\em pre-trained} network $\pmb{\epsilon}_{\pmb{\theta}}$  (trained with the standard algorithm Alg.~\ref{alg:1}) to run the reverse diffusion for $t$ steps. Note that  we adopt the equation in line 4 of Alg.~\ref{alg:2} but we remove the stochastic term $\sigma_t \pmb{z}$. 
 Differently from the analogous experiment presented in Sec.~\ref{sec.Exposure},
 this deterministic reverse diffusion process allows the model to target the mode of $\pmb{x}_{0}$ instead of favouring diversity \cite{UnderstandingDM}. Finally, we use the average pixel-wise $L_1$ distance between $\pmb{x}_{0}$ and $\hat{\pmb{x}}_{0}$ to estimate the cumulative error computed in the whole trajectory of $t$ steps. Note that, since each pixel is normalized in $[-1, 1]$ (Sec.~\ref{sec.Background}), then this distance is upper bounded by 2.

\begin{algorithm}[h]
   \caption{Deterministic measurement of exposure bias}
   \label{alg:3}
   \begin{algorithmic}[1]
        \STATE Initialize $\delta_t = 0$, $n_t = 0$ ($\forall t \in \{1, ..., T \}$)
        \REPEAT
        \STATE $\pmb{x}_0 \sim q(\pmb{x}_0)$, $t\sim \mathbb{U}(\{1,...,T\})$, $\pmb{\epsilon} \sim {\cal N} (\pmb{0}, \pmb{I})$
        \STATE compute  $\pmb{x}_t$ using Eq.~\ref{eq.x-t}
        
        \FOR{$\tau := t, ..., 1$}
        \STATE $\hat{\pmb{x}}_{\tau-1} = \frac{1}{\sqrt{\alpha_{\tau}}} (\hat{\pmb{x}}_{\tau} - \frac{1-\alpha_{\tau}}{\sqrt{1-\bar{\alpha}_{\tau}}} \pmb{\epsilon}_{\pmb{\theta}} (\hat{\pmb{x}}_{\tau}, \tau))$ 
        \ENDFOR
        \STATE $\delta_t = \delta_t + ||\pmb{x}_{0} - \hat{\pmb{x}}_{0}||_1/M$, where $M$ is the number of pixels in $\pmb{x}_{0}$
        \STATE $n_t = n_t + 1$
        \UNTIL {$N$ iterations}
        \STATE if $n_t \neq 0$, then  $\bar{\delta _{t}} = \frac{\delta_t}{n_t}$ ($\forall t \in \{1, ..., T \}$)
    \end{algorithmic}
\end{algorithm}

\begin{figure}[h]
    \centering
    \includegraphics[scale=0.4]{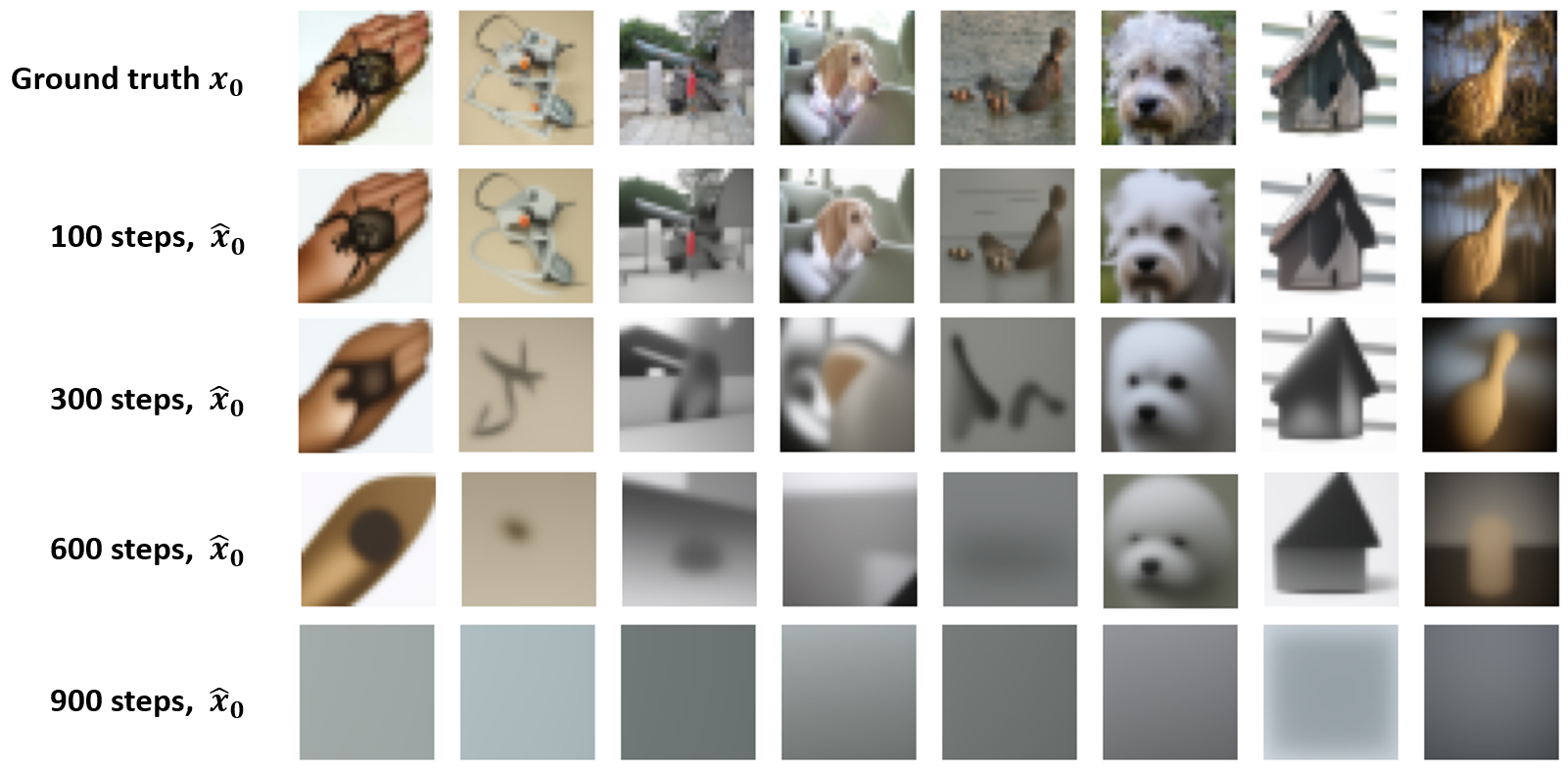}
    \caption{Visualization of the exposure bias problem with different diffusion chain lengths. 
    }
    \label{exposure bias}
\end{figure}

In Tab.~\ref{deterministic exposure bias},  we
report the  exposure bias measured using $\bar{\delta _{t}}$ with respect to different trajectory lengths ($t$). This table shows that the error accumulates greatly as the number of reverse diffusion steps increases. 
 In Fig.~\ref{exposure bias} 
 we visualize a few pairs of images $(\pmb{x}_{0}, \hat{\pmb{x}}_{0})$ with the corresponding length of the diffusion trajectory ($t$). 
 These images clearly show how large the error is accumulated with the diffusion chain getting longer.

\begin{table}[h]
\caption{A deterministic estimate  of the exposure bias ($\bar{\delta _{t}}$) with respect to different lengths of the reverse diffusion  trajectory. The error is upper bounded by 2.
}
\label{deterministic exposure bias}
\begin{center}
\begin{tabular}{@{}lllll@{}}
\toprule
\multirow{2}{*}{Model} & \multicolumn{4}{l}{Number of reverse diffusion steps} \\ \cmidrule(l){2-5} 
 & 100 & 300 & 600 & 1,000 \\ \midrule
ADM & 0.0539 & 0.1074 & 0.1821 & 0.8165 \\
%ADM-IP & 0.0538 & 0.1070 & 0.1800 & 0.7702 \\ 
\bottomrule
\end{tabular}
\end{center}
\end{table}

\subsection{Distribution of the Perturbed Input}
\label{sec:y-distributio-proof}

In this section, we prove that $\pmb{y}_t$ is Gaussian distributed as described in Eq.~\ref{eq.y-distribution}.
Generally speaking, if $A \sim  {\cal N} (\mu_A, \sigma_A^2)$
and $B \sim  {\cal N} (\mu_B, \sigma_B^2)$ are two independent Gaussian distributed random variables, then 
its linear combination $S = aA + bB$ (with $a,b$ two scalars) is also Gaussian distributed:

\begin{equation}
\label{eq.generic-linear-comb-rand-var}
S \sim  {\cal N} (a \mu_A + b \mu_B, a^2 \sigma_A^2 + b^2 \sigma_B^2).
\end{equation}

In our case, we have that, for a given $\pmb{x}_0$, $\pmb{y}_t$ is a linear combination of $\pmb{x}_t$ and $\pmb{\xi}$, which are two independent, Gaussian distributed random variables:

\begin{eqnarray}
\label{eq.proof-intermed-steps}
q(\pmb{x}_t | \pmb{x}_0) =  {\cal N} (\pmb{x}_t; \sqrt{\bar{\alpha}_t} \pmb{x}_0, (1 - \bar{\alpha}_t) \pmb{I}), \\
\pmb{\xi} \sim {\cal N} (\pmb{0}, \pmb{I}), \\
\pmb{y}_t = \pmb{x}_t +   \sqrt{1 - \bar{\alpha}_t}  \gamma \pmb{\xi}. %\label{eq.y-t-computed-from-x-t}
\end{eqnarray}

\noindent
Hence, if in Eq.~\ref{eq.generic-linear-comb-rand-var} we replace $S$ with $\pmb{y}_t$, $A$ with $\pmb{x}_t$, $B$ with $\pmb{\xi}$, and we use $a = 1$ and $b = \sqrt{1 - \bar{\alpha}_t}  \gamma$, we get:

\begin{eqnarray}
\label{eq.y-distribution-proof}
q(\pmb{y}_t | \pmb{x}_0) =  {\cal N} (\pmb{y}_t; \sqrt{\bar{\alpha}_t} \pmb{x}_0, (1 - \bar{\alpha}_t) \pmb{I} + \gamma^2 (1 - \bar{\alpha}_t) \pmb{I}) = \\
=  {\cal N} (\pmb{y}_t; \sqrt{\bar{\alpha}_t} \pmb{x}_0, (1 - \bar{\alpha}_t) (1 + \gamma^2) \pmb{I}).
\end{eqnarray}

\subsection{Ablation Study: Input Perturbation is not Equivalent to Using a Different Noise Variance}
\label{sec:Ablation study}

The goal of this section is to empirically show that  DDPM-IP is not equivalent to using a standard DDPM algorithm with a different noise distribution. Following the discussion in Sec.~\ref{sec.Discussion}, and adopting the same terminology, we compare DDPM-IP with DDPM-$y$, where the latter is trained using the standard algorithm (Alg.~\ref{alg:1}) but adopting the noise distribution of $\pmb{y}_t$. 
Tab.~\ref{Input perturbation and New Noise Factor} shows that DDPM-$y$ is even worse than DDPM.

\begin{table}[H]
\caption{CIFAR10: comparing DDPM, DDPM-$y$ and DDPM-IP using different numbers of revers diffusion  steps.}

\label{Input perturbation and New Noise Factor}
\begin{center}
\begin{tabular}{@{}lllllllllll@{}}
\toprule
\multirow{2}{*}{Model} & \multirow{2}{*}{Input} & \multirow{2}{*}{Target} & \multicolumn{2}{l}{80 steps} & \multicolumn{2}{l}{100 steps} & \multicolumn{2}{l}{300 steps} & \multicolumn{2}{l}{1000 steps} \\ \cmidrule(lr){4-5} \cmidrule(lr){6-7} \cmidrule(lr){8-9} \cmidrule(lr){10-11} 
 &  &  & FID & sFID & FID & sFID & FID & sFID & FID & sFID \\ \midrule
DDPM & $\pmb{x}_t$ & $\pmb{\epsilon}$ & 3.63 & 5.97 & 3.37 & 5.66 & 2.95 & 4.95 & 2.99 & 4.76 \\
DDPM-$y$ & $\pmb{y}_t$ & $\pmb{\epsilon^\prime}$ & 4.24 & 6.51 & 3.90 & 6.23 & 3.21 & 5.39 & 3.25 & 5.04 \\
DDPM-IP & $\pmb{y}_t$ & $\pmb{\epsilon}$ & \textbf{2.93} & \textbf{4.69} & \textbf{2.70} & \textbf{4.51} & \textbf{2.67} & \textbf{4.14} & \textbf{2.76} & \textbf{4.05} \\ \bottomrule
\end{tabular}
\end{center}
\end{table}

\subsection{Recall and Precision}
\label{sec:recall and precision}

We compare Recall and Precision for ADM and ADM-IP using the improved metrics \citep{kynkaanniemi2019improved} and the code of \citet{ADM}. For each dataset and model, we generate 50,000 samples with 1,000 sampling steps. The results in Tab. \ref{recall and precision compare} indicate that the Recall and Precision values achieved by ADM and ADM-IP have no significant difference, while ADM-IP gets slightly better results on CIFAR10 32$\times$32 and FFHQ 128$\times$128. Note that, due to the limited memory of our NVIDIA V100 16G GPU, we experienced an out-of-memory issue when computing  Recall and Precision on the ImageNet 32$\times$32 dataset, thus this result is not reported in Tab. \ref{recall and precision compare}.    

\begin{table}[H]
\caption{Comparing  Recall and Precision for ADM and ADM-IP on the four datasets using 1,000 sampling steps.}
\label{recall and precision compare}
\begin{center}
\begin{tabular}{@{}llcccllll@{}}
\toprule
\multirow{2}{*}{Model} & \multicolumn{2}{l}{CIFAR10 32$\times$32} & \multicolumn{2}{l}{LSUN tower 64$\times$64} & \multicolumn{2}{l}{CelebA 64$\times$64} & \multicolumn{2}{l}{FFHQ 128$\times$128} \\ \cmidrule(lr){2-3} \cmidrule(lr){4-5} \cmidrule(lr){6-7} \cmidrule(lr){8-9}
 & \multicolumn{1}{c}{Recall} & Precision & Recall & Precision & \multicolumn{1}{c}{Recall} & \multicolumn{1}{c}{Precision} & \multicolumn{1}{c}{Recall} & \multicolumn{1}{c}{Precision} \\ \midrule
ADM & 0.600 & 0.690 & \textbf{0.618} & 0.631 & 0.592 & \textbf{0.703} & 0.583 & 0.690 \\
ADM-IP & \textbf{0.606} & \textbf{0.696} & 0.612 & \textbf{0.640} & \textbf{0.601} & 0.700 & \textbf{0.585} & \textbf{0.703} \\ \bottomrule
\end{tabular}
\end{center}
\end{table}

\subsection{Gaussian Prediction Error}
\label{sec:gaussian prediction error}

\begin{figure}[h]
\centering
        \subfigure[$e_{t}^{i}$ with $t=300, i=1$]{
		\begin{minipage}{5cm}
		\includegraphics[width=\textwidth]{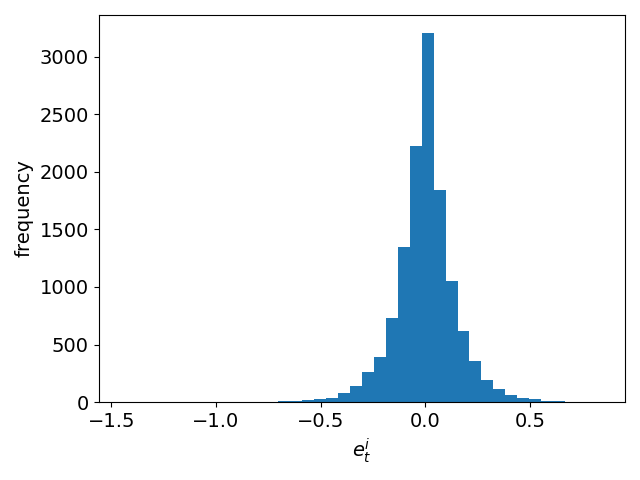}
		\label{pixel1}
		\end{minipage}
		\hspace{0mm}
        }
        \subfigure[$e_{t}^{i}$ with $t=300, i=1025$]{
		\begin{minipage}{5cm}
		\includegraphics[width=\textwidth]{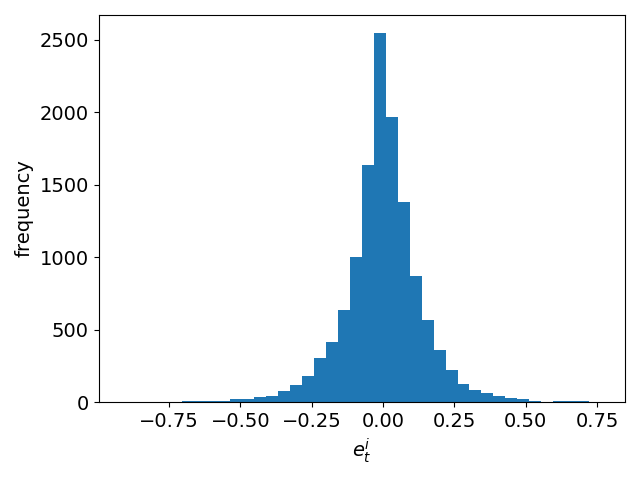}
		\label{pixel1025}
		\end{minipage}
		\hspace{0mm}
        }
        \subfigure[$e_{t}^{i}$ with $t=300, i=2049$]{
		\begin{minipage}{5cm}
		\includegraphics[width=\textwidth]{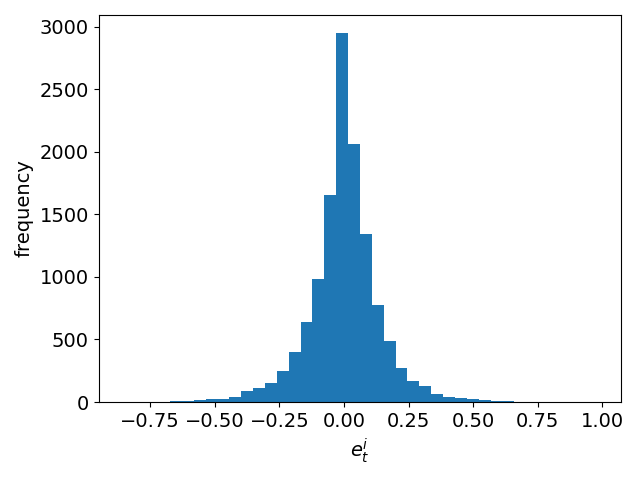}
		\label{pixel2049}
		\end{minipage}
		\hspace{0mm}
        }
  
        \subfigure[$e_{t}^{i}$ with $t=600, i=528$]{
		\begin{minipage}{5cm}
		\includegraphics[width=\textwidth]{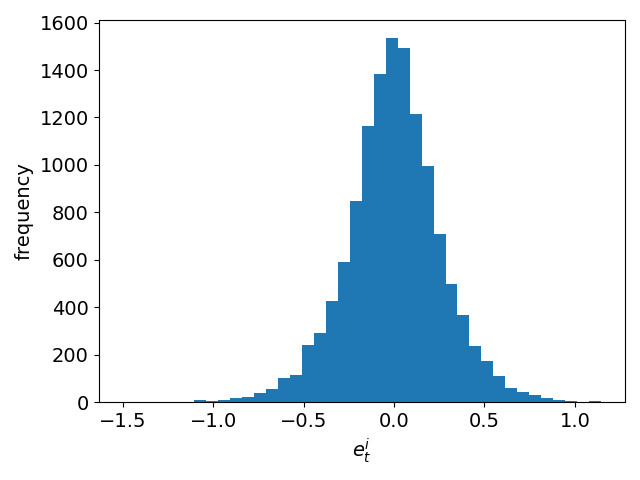}
		\label{pixel528}
		\end{minipage}
		\hspace{0mm}
        }
        \subfigure[$e_{t}^{i}$ with $t=600, i=1552$]{
		\begin{minipage}{5cm}
		\includegraphics[width=\textwidth]{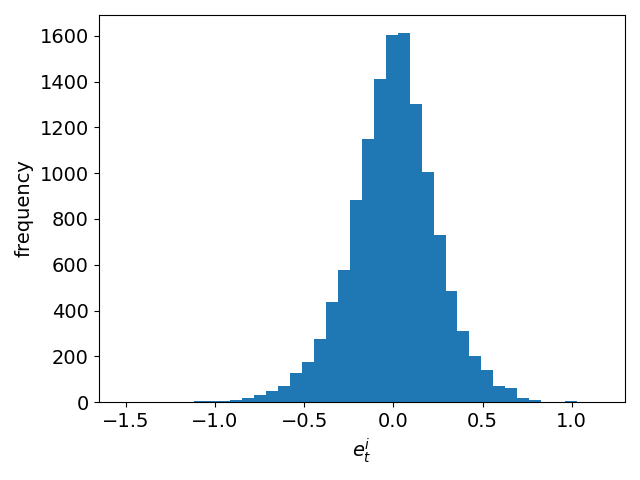}
		\label{pixel1552}
		\end{minipage}
		\hspace{0mm}
        }
        \subfigure[$e_{t}^{i}$ with $t=600, i=2576$]{
		\begin{minipage}{5cm}
		\includegraphics[width=\textwidth]{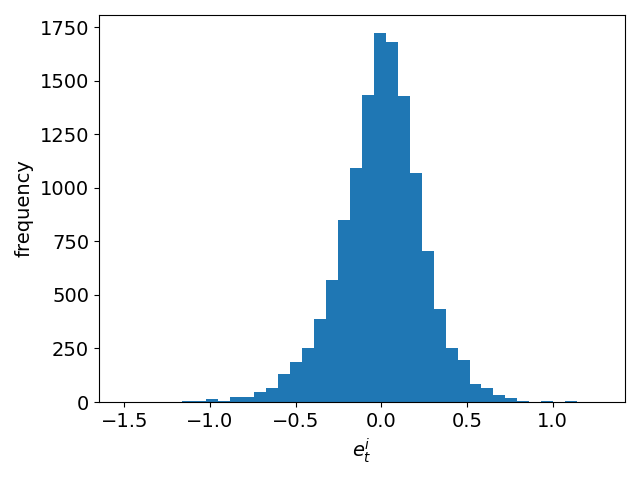}
		\label{pixel2576}
		\end{minipage}
		\hspace{0mm}
        }
  
	\subfigure[$e_{t}^{i}$ with $t=900, i=1024$]{
		\begin{minipage}{5cm}
         \includegraphics[width=\textwidth]{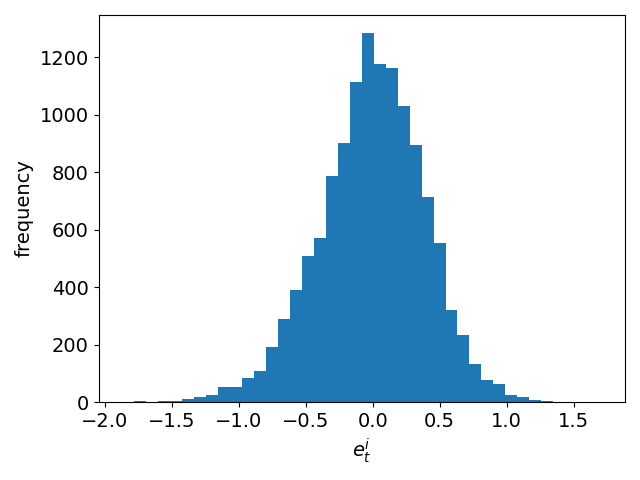} 
         \label{pixel1024}
		\end{minipage}
		\hspace{0mm}
	}
	\subfigure[$e_{t}^{i}$ with $t=900, i=2048$]{
		\begin{minipage}{5cm}
		\includegraphics[width=\textwidth]{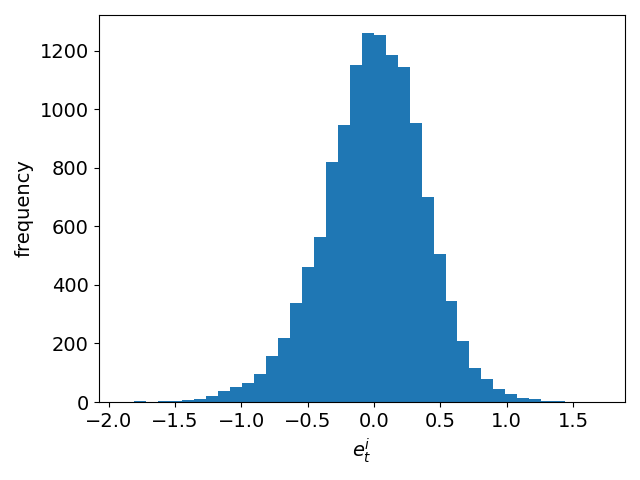}
		\label{pixel2048}
		\end{minipage}
		\hspace{0mm}
	}
	\subfigure[$e_{t}^{i}$ with $t=900, i=3072$]{
		\begin{minipage}{5cm}
		\includegraphics[width=\textwidth]{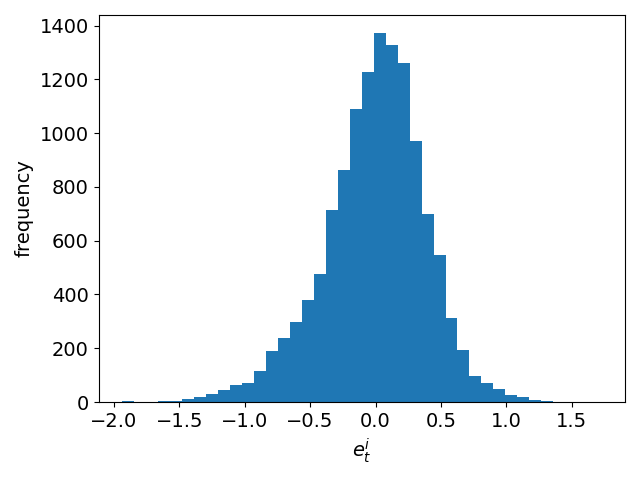}
		\label{pixel3072}
		\end{minipage}
		\hspace{0mm}
	}
\caption{The empirical distribution of $e_{t}^{i}$ with different random values of $t$ and $i$.} 
\label{et histogram}
\end{figure}

In this section, we use ImageNet 32$\times$32 to empirically show that $\pmb{e}_{t} \sim {\cal N} (\pmb{0}, \nu_{t}^{2}\pmb{I})$
(Sec.~\ref{subsec.gamma choice}), i.e., that 
the prediction error is nearly isotropic Gaussian distributed.
To do so, we need to prove that, for each $t$ and each input dimension (i.e., for each pixel and color channel) $i \in \{1, ..., M \}$, 
the pixel-wise error ($e_t^i$) follows $e_t^i \sim {\cal N} (0, \nu_{t}^{2})$.
To test this hypothesis,
 we uniformly select a subset of 100 $t$ values in $\{1, ..., T\}$ using a stride of 10. 
Then, for each $t$,  we use 10K images and all the pixels to compute the pixel independent mean $\mu_t$ and variance $\nu_t^2$ of the error, which we use to standardize the error values for all the pixels $e_t^i$
(i.e., $\bar{e}_t^i = \frac{e_t^i - \mu_t}{\nu_t^2}$).
Then, 
for each $i$, we use 50 randomly selected $\bar{e}_t^i$ values
and the Shapiro–Wilk test \cite{shapiro1965analysis} 
to verify that they follow a standard normal distribution.
 The confidence level is set at 95\% and we reject the null hypothesis if the p-value is less than 0.05. 
 The null hypothesis was  rejected only in a small minority of cases, confirming that the 
 error $\pmb{e}_{t}$ is almost isotropic Gaussian distributed.
 Fig.~\ref{et histogram} shows a few histogram examples for $e_{t}^{i}$ computed at different pixels.

\subsection{Relation between the Lipschitz Constant Minimization  and the Weight Decay Minimization}
\label{Lips and weight decay}

By definition, in Lipschitz continuos functions, the relation between the  
  output difference $\left\| f_{w}(x_{1})-f_{w}(x_{2}) \right\|$ and the input difference $\left\| x_{1}-x_{2} \right\|$ of two points is governed by a constant $K$ as follows:

\begin{equation}
    \label{eq.proof1}
\left\| f_{w}(x_{1})-f_{w}(x_{2}) \right\| \leq K\cdot \left\| x_{1}-x_{2} \right\|.
\end{equation}

Since a neural network is usually a stack of  layers, without loss of generality we consider a single layer neural network, $f(x) = ReLU(Wx+b)$, thus we have:
\begin{equation}
    \label{eq.proof2}
\left\| f(Wx_{1}+b) - f(Wx_{2}+b) \right\| \leq K\cdot \left\| x_{1}-x_{2} \right\|.
\end{equation}

Using the first order term of Tylor Series to approximate the left side of the above equation, we get:
\begin{equation}
    \label{eq.proof3}
\left\| \frac{\partial f}{\partial y}\cdot W(x_{1}-x_{2}) \right\| \leq K\cdot \left\|x_{1}-x_{2} \right\|,
\end{equation}

where the details of Tylor Series approximation are:
\begin{itemize}
\item Let $y=Wx+b$, we approximate $f(y)$ at the point $y=0$.
\item Hence, $f(y)\approx f(0)+f'(0)(y-0)f(y)\approx f(0)+f'(0)(y-0)$.
\item Substitute $y$ with $y_{1}$ and $y_{2}$, where $y_{1}=Wx_{1}+b, y_{2}=Wx_{2}+b$.
\item Thus, $f(y_{1})-f(y_{2})\approx f(0)+f'(0)y_{1}-f(0)-f'(0)y_{2}=f'(0)(y_{1}-y{2})=f'(0)W(x_{1}-x_{2})$.
\end{itemize}

Since $\frac{\partial f}{\partial y}$ is bounded by 1 when $f=ReLU$, we can ignore it, and we have:

\begin{equation}
    \label{eq.proof4}
\left\| W(x_{1}-x_{2}) \right\|\leq K\left\| x_{1}-x_{2}\right\|.
\end{equation}
We now  introduce the Spectral Norm $\left\| W \right\|_{2}$. According to the definition $\left\| W \right\|_{2}=\displaystyle \max_{x\neq0}\frac{\left\|Wx\right\|}{\left\| x\right\|}$ , we have:

\begin{equation}
    \label{eq.proof5}
\left\| W(x_{1}-x_{2}) \right\| \leq \left\|W\right\|_{2}\cdot \left\| x_{1}-x_{2} \right\|.
\end{equation}

Comparing Eq.~\ref{eq.proof4} with Eq.~\ref{eq.proof5}, we can use $\left\| W \right\|_{2}$ as the Lipschitz constant $K$. 
We can use the Frobenius Norm $\left\| W \right\|_{F}$ to approximate the Spectral Norm $\left\| W \right\|_{2}$ because, using the Cauchy inequality, we have:
 
\begin{equation}
    \label{eq.proof6}
\left\| Wx \right\| \leq \left\|W\right\|_{F}\cdot \left\| x \right\|,
\end{equation}

where the definition of the Frobenius Norm is: $\left\| W \right\|_{F}=\sqrt{\sum_{i,j}w_{i,j}^{2}}$.

Thus, we can use the Frobenius Norm $\left\| W \right\|_{F}$ to approximate the constant $K$. Minimizing this constant during training is often implemented by adding a loss term $\lambda \left\| W \right\|_{F}^{2}$ to the loss function. This loss term is exactly the Weight Decay according to the definition of $\left\| W \right\|_{F}=\sqrt{\sum_{i,j}w_{i,j}^{2}}$.

\subsection{Hyperparameters}
\label{sec:hyperparameters}

For both ADM and ADM-IP, we use the hyperparameters specified in \cite{ADM}, except for LSUN tower, for which we used a  resolution of 64$\times$64. The hyperparameter values are reported in Tab.~\ref{ADM hyperparameters}. We train all the  models using the AdamW optimizer \cite{AdamW}. Furthermore, we use 16-bit precision and loss-scaling \cite{loss_scaling}  for mixed precision training, but keeping 32-bit weights, EMA, and the optimizer state. We use an EMA rate of 0.9999 for all the experiments. These settings are the same as in \cite{ADM}.

We use Pytorch 1.8 \cite{pytorch} and trained all the models on different NVIDIA Tesla V100s (16G memory). In more detail, we use 2 GPUs to train the models on CIFAR10 for 2 days, and 4 GPUs to train the models on ImageNet 32$\times$32 for 34 days. For LSUN tower 64$\times$64, CelebA 64$\times$64 and FFHQ 128$\times$128, we used 16 GPUs to train the models for 3 days, 5 days and 4 days, respectively.

\begin{table*}[h]
\caption{ADM and ADM-IP hyperparameter values}
\label{ADM hyperparameters}
\begin{center}
\begin{tabular}{@{}llllll@{}}
\toprule
 & \begin{tabular}[c]{@{}l@{}}CIFAR10 \\ 32$\times$32\end{tabular} & \begin{tabular}[c]{@{}l@{}}ImageNet \\ 32$\times$32\end{tabular} & \begin{tabular}[c]{@{}l@{}}LSUN tower \\ 64$\times$64\end{tabular} & \begin{tabular}[c]{@{}l@{}}CelebA \\ 64$\times$64\end{tabular} & \begin{tabular}[c]{@{}l@{}}FFHQ \\ 128$\times$128\end{tabular}  \\ \midrule
Diffusion steps & 1,000 & 1,000 & 1,000 & 1,000 & 1,000 \\
Noise schedule & cosine & cosine & cosine & cosine & cosine \\
Model size & 57M & 57M & 295M & 295M & 543M \\
Channels & 128 & 128 & 192 & 192 & 256 \\
Residual blocks & 3 & 3 & 3 & 3 & 3 \\
Channels multiple & 1, 2, 2, 2 & 1, 2, 2, 2 & 1, 2, 3, 4 & 1, 2, 3, 4 & 1, 1, 2, 3, 4 \\
Heads channels & 32 & 32 & 64 & 64 & 64 \\
Attention resolution & 16, 8 & 16, 8 & 32, 16, 8 & 32, 16, 8 & 32, 16, 8 \\
BigGAN up/downsample & True & True & True & True & True \\
Dropout & 0.3 & 0.3 & 0.1 & 0.1 & 0.1 \\
Batch size & 128 & 512 & 256 & 256 & 128 \\
Training iterations & 540K & 5000K & 340K & 540K & 480K \\
Training images & 50K & 1281K & 708K & 203K & 70K \\
Learning rate & 1e-4 & 1e-4 & 1e-4 & 1e-4 & 1e-4 \\ \bottomrule
\end{tabular}
\end{center}
\end{table*}

Regarding the DDIM and DDIM-IP experiments, we use the default hyperparameters specified in the public code  of \citet{DBLP:conf/iclr/SongME21}. We train both DDIM and DDIM-IP on CIFAR10 from scratch for 1600K iterations with batch size 128. The complete list of hyperparameters is shown in Tab.~\ref{DDIM hyperparameters}. We train DDPM/DDPM-IP with a single NVIDIA Tesla V100s (16G memory) for 8 days on a Pytorch 1.8 platform.

\begin{table*}[h]
\caption{DDIM and DDIM-IP hyperparameter values on CIFAR10 dataset}
\label{DDIM hyperparameters}
\begin{center}
\begin{tabular}{@{}ll|ll@{}}
\toprule
Diffusion Steps      & 1000    & Variance type   & fixed large \\
Noise schedule       & linear  & Ema rate        & 0.9999      \\
Channels             & 128     & Batch size      & 128         \\
Channels multiple    & 1,2,2,2 & Iterations      & 1600K       \\
Residual blocks      & 2       & Training images & 50K         \\
Attention resolution & 16      & Optimizer       & Adam        \\
Dropout              & 0.1     & Learning rate   & 2e-4        \\ \bottomrule
\end{tabular}
\end{center}
\end{table*}

\subsection{Qualitative Comparison between ADM and ADM-IP}
\label{sec:qualitative results comparison}

\begin{figure}[h]
\centering
	\subfigure[Samples generated by ADM trained on CIFAR10 (FID 2.99)]{
        \hspace{10mm}
		\begin{minipage}{8cm}
         \includegraphics[width=\textwidth]{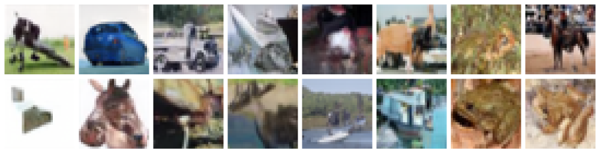} 
         \label{cifar base}
		\end{minipage}
		\hspace{10mm}
	}
	\subfigure[Samples generated by ADM-IP trained on CIFAR10 (FID 2.76)]{
        \hspace{10mm}
		\begin{minipage}{8cm}
		\includegraphics[width=\textwidth]{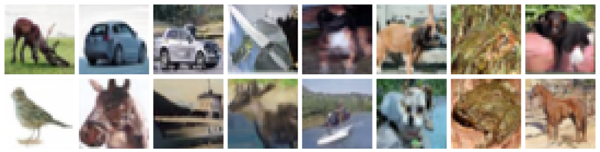}
		\label{cifar noise}
		\end{minipage}
		\hspace{10mm}
	}
\caption{CIFAR10, qualitative results. The samples are generated using 1,000 sampling steps.} 
\label{sample comparison cifar10}
\end{figure}

\begin{figure}[h]
\centering
	\subfigure[Samples generated by ADM trained on ImageNet 32$\times$32 (FID 3.53)]{
        \hspace{10mm}
		\begin{minipage}{8cm}
         \includegraphics[width=\textwidth]{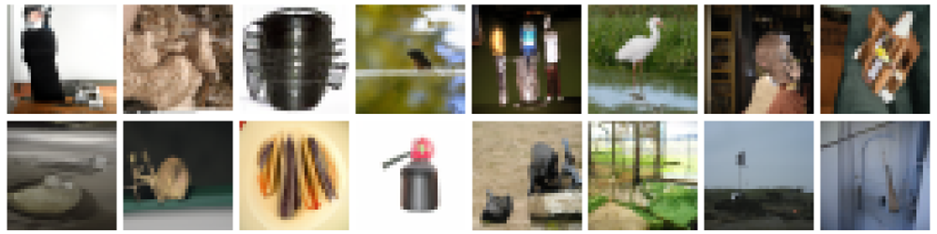} 
         \label{imagenet32 base}
		\end{minipage}
		\hspace{10mm}
	}
	\subfigure[Samples generated by ADM-IP trained on ImageNet 32$\times$32 (FID 2.72)]{
        \hspace{10mm}
		\begin{minipage}{8cm}
		\includegraphics[width=\textwidth]{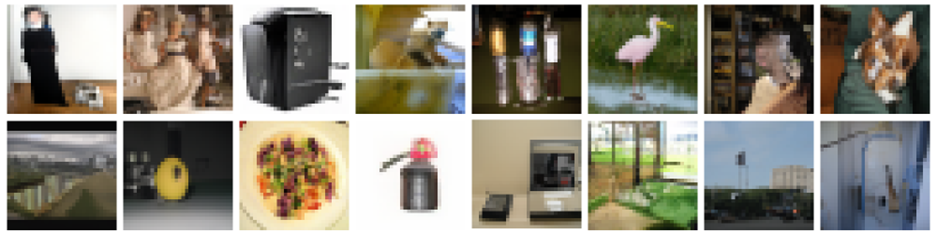}
		\label{imagenet32 noise}
		\end{minipage}
		\hspace{10mm}
	}
\caption{ImageNet 32$\times$32, qualitative results. The samples are generated using 1,000 sampling steps.} 
\label{sample comparison imagenet32}
\end{figure}

\begin{figure}[h]
\centering
	\subfigure[Samples generated by ADM trained on LSUN tower 64$\times$64 (FID 3.39)]{
		\begin{minipage}{9cm}
         \includegraphics[width=\textwidth]{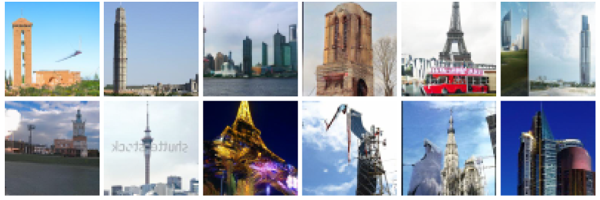} 
         \label{lsun base}
		\end{minipage}
		\hspace{0mm}
	}
	\subfigure[Samples generated by ADM-IP trained on LSUN tower 64$\times$64 (FID 2.68)]{
		\begin{minipage}{9cm}
		\includegraphics[width=\textwidth]{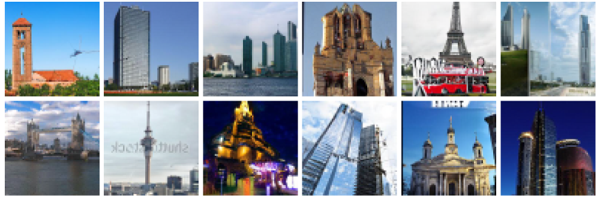}
		\label{lsun noise}
		\end{minipage}
		\hspace{0mm}
	}
\caption{LSUN tower 64$\times$64, qualitative results.  The samples are generated using 1,000 sampling steps.} 
\label{sample comparison lsun}
\end{figure}

\begin{figure}[h]
\centering
	\subfigure[Samples generated by ADM trained on CelebA 64$\times$64 (FID 1.60)]{
		\begin{minipage}{9cm}
         \includegraphics[width=\textwidth]{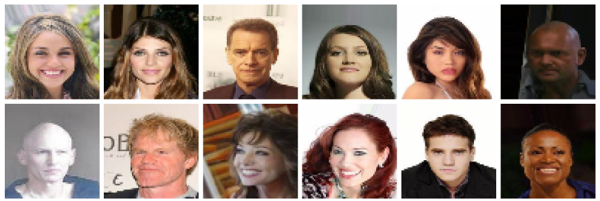} 
         \label{celeba base}
		\end{minipage}
		\hspace{0mm}
	}
	\subfigure[Samples generated by ADM-IP trained on CelebA 64$\times$64 (FID 1.31)]{
		\begin{minipage}{9cm}
		\includegraphics[width=\textwidth]{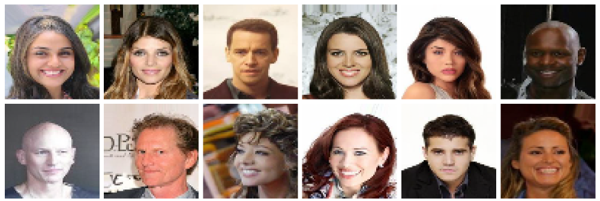}
		\label{celeba noise}
		\end{minipage}
		\hspace{0mm}
	}
\caption{CelebA 64$\times$64, qualitative results.  The samples are generated using 1,000 sampling steps.} 
\label{sample comparison celeba}
\end{figure}

\begin{figure}[h]
\centering
	\subfigure[Samples generated by ADM trained on FFHQ 128$\times$128 (FID 9.65)]{
		\begin{minipage}{13cm}
         \includegraphics[width=\textwidth]{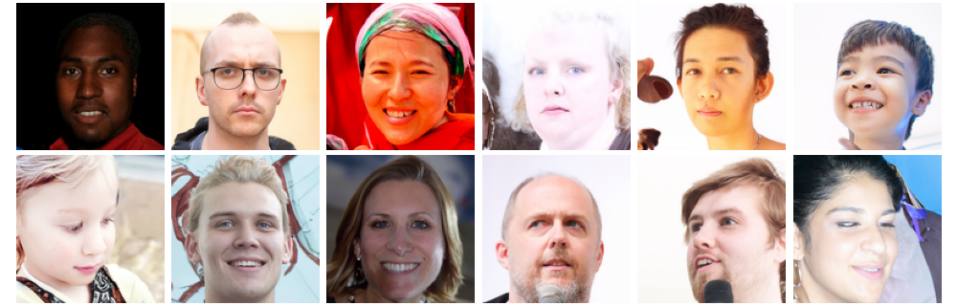} 
         \label{ffhq base}
		\end{minipage}
		\hspace{0mm}
	}
	\subfigure[Samples generated by ADM-IP trained on FFHQ 128$\times$128 (FID 2.98)]{
		\begin{minipage}{13cm}
		\includegraphics[width=\textwidth]{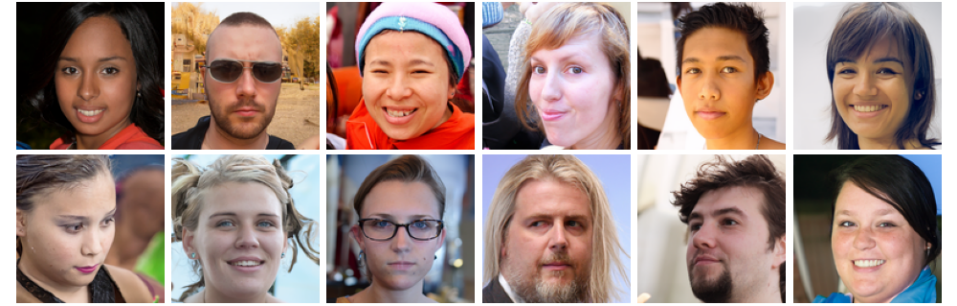}
		\label{ffhq noise}
		\end{minipage}
		\hspace{0mm}
	}
\caption{FFHQ 128$\times$128, qualitative results.  The samples are generated using 1,000 sampling steps.} 
\label{sample comparison ffhq}
\end{figure}

In this section, we qualitatively compare ADM with ADM-IP. For a fair comparison, we start sampling  the same $\pmb{x}_T$ for both models. Fig.~\ref{sample comparison cifar10}, \ref{sample comparison imagenet32}, \ref{sample comparison lsun}, \ref{sample comparison celeba}, \ref{sample comparison ffhq} show  that the images generated by ADM-IP are usually comparable  or better  than those produced by ADM. For example, in Fig.~\ref{sample comparison cifar10}, ADM fails to run into the bird, the boat and the dog modes in the first, the third and the sixth image on the second row. Similarly, in Fig.~\ref{sample comparison lsun}, ADM fails to complete the  building in the fourth image on the second row. Moreover, the details and colors of the towers generated by ADM-IP are more visually realistic and appealing. Finally, on the FFHQ 128$\times$128 dataset, the ADM generated samples   suffer from overexposure and loss of background detail, whereas the ADM-IP samples   do not (see Fig.~\ref{sample comparison ffhq}).

\subsection{Additional Qualitative Results for ADM-IP}
\label{sec:qualitative results}
\begin{figure}[h]
    \centering
    \includegraphics[scale=0.5]{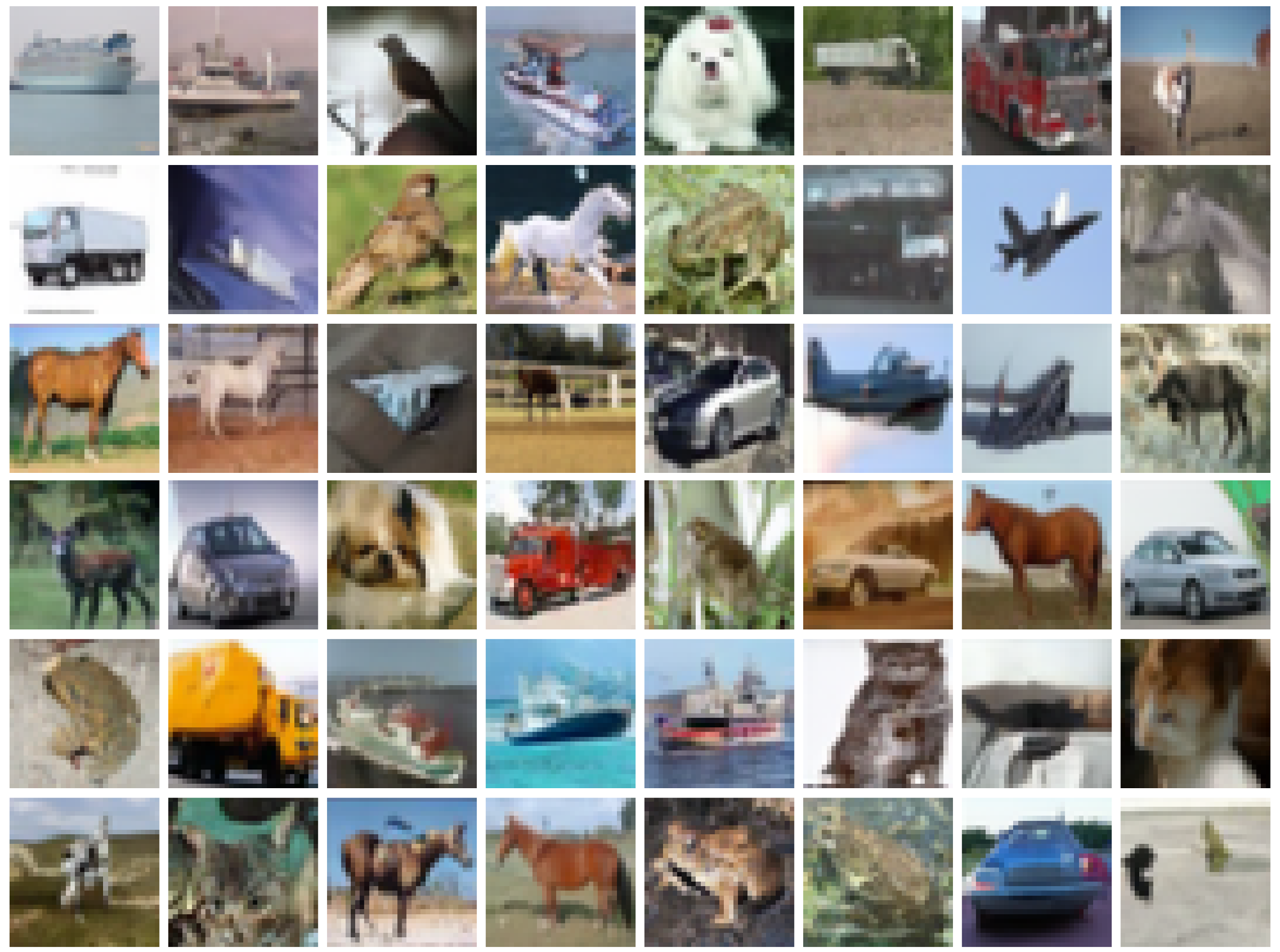}
    \caption{Samples generated by ADM-IP trained on CIFAR10  (FID 2.67 , 300 sampling steps)}
    \label{cifar10 samples}
\end{figure}

\begin{figure}[h]
    \centering
    \includegraphics[scale=0.5]{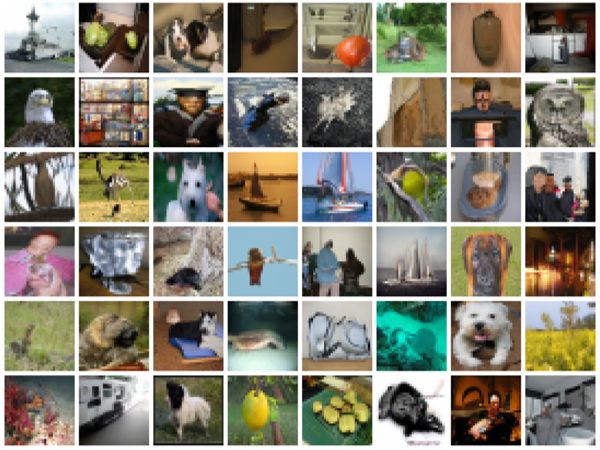}
    \caption{Samples generated by ADM-IP trained on ImageNet 32$\times$32  (FID 2.66, 300 sampling steps)}
    \label{imagenet32 samples}
\end{figure}

\begin{figure}[h]
    \centering
    \includegraphics[scale=0.57]{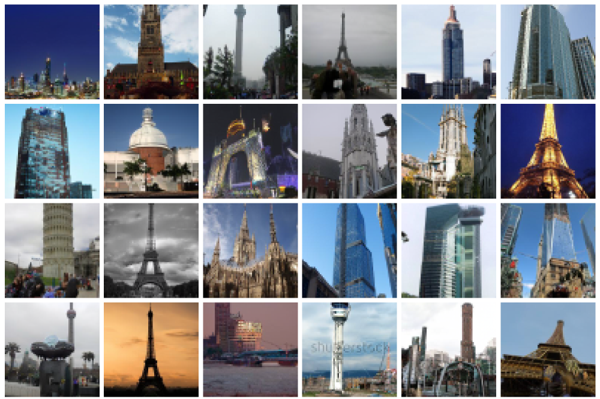}
    \caption{Samples generated by ADM-IP trained on LSUN tower 64$\times$64  (FID 2.60 , 300 sampling steps)}
    \label{LSUN tower samples}
\end{figure}

\begin{figure}[h]
    \centering
    \includegraphics[scale=0.57]{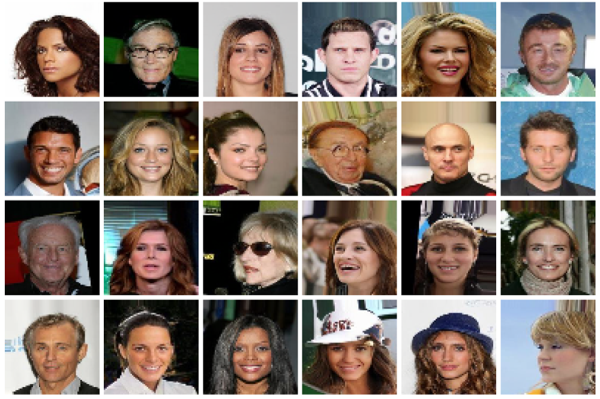}
    \caption{Samples generated by ADM-IP trained on CelebA 64$\times$64  (FID 1.27, 900 sampling steps)}
    \label{celeba samples}
\end{figure}

\begin{figure}[h]
    \centering
    \includegraphics[scale=0.5]{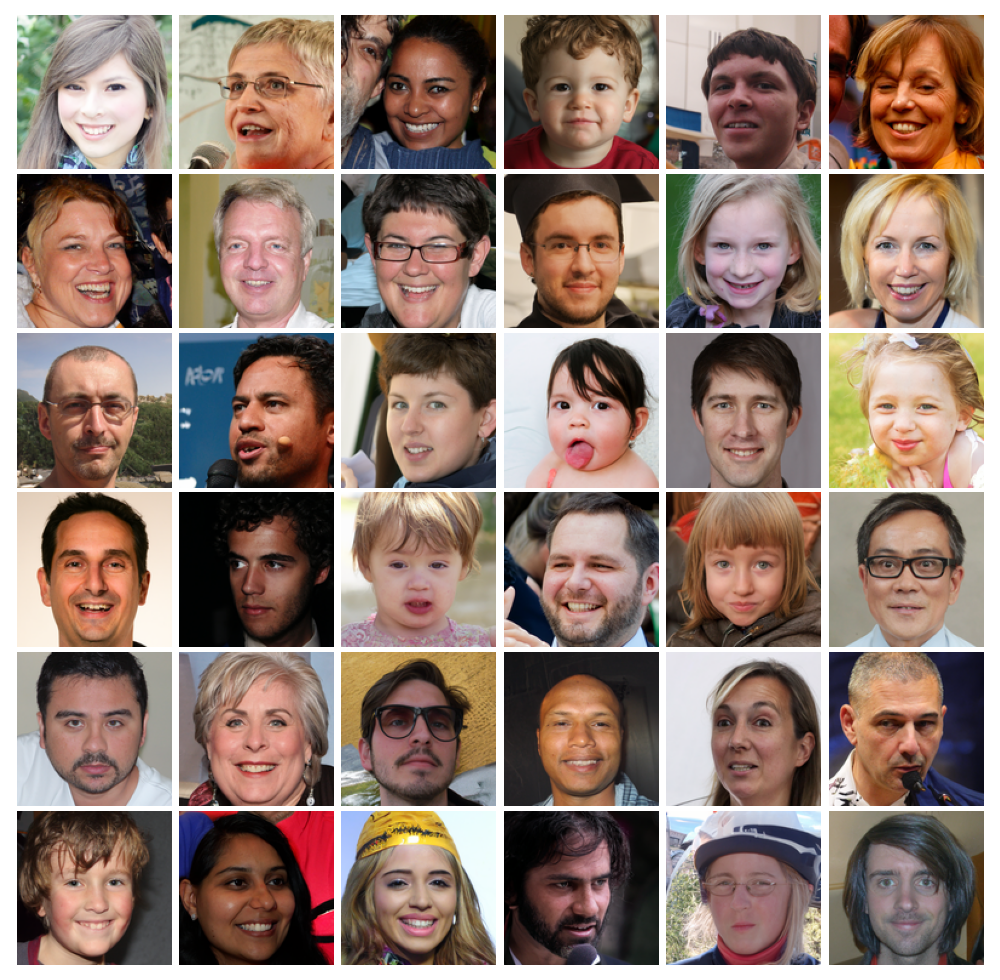}
    \caption{Samples generated by ADM-IP trained on FFHQ 128$\times$128  (FID 2.98, 1,000 sampling steps)}
    \label{ffhq samples}
\end{figure}

We show additional images generated by our ADM-IP models trained on CIFAR10 (Fig.~\ref{cifar10 samples}), ImageNet 32$\times$32 (Fig.~\ref{imagenet32 samples}), LSUN tower 64$\times$64 (Fig.~\ref{LSUN tower samples}), CelebA 64$\times$64 (Fig.~\ref{celeba samples}) and FFHQ 128$\times$128 (Fig.~\ref{ffhq samples}). 
For each dataset, we used the  best number of sampling steps as indicated in  Tab.~\ref{DDPM results}. 

\end{document}